\documentclass[runningheads]{llncs}
\usepackage{authblk}
\usepackage{graphicx}
\usepackage{tikz}
\usepackage{comment}
\usepackage{amsmath,amssymb} 

\usepackage{color}

\usepackage{hyperref}
\usepackage{url}
\usepackage{multirow}
\usepackage{graphicx}
\usepackage[T1]{fontenc}
\usepackage[latin9]{inputenc}
\usepackage{array}
\usepackage{subfigure}
\usepackage[symbol*]{footmisc}
\DefineFNsymbolsTM{myfnsymbols}{
  \textasteriskcentered *
  \textdagger    \dagger
  \textdaggerdbl \ddagger
  \textsection   \mathsection
  \textbardbl    \|%
  \textparagraph \mathparagraph
}%
\setfnsymbol{myfnsymbols}
\usepackage{multirow}
\usepackage{booktabs}
\usepackage{color}

\makeatletter
\newcommand{\printfnsymbol}[1]{%
  \textsuperscript{\@fnsymbol{#1}}%
}
\makeatother
\begin{document}
\pagestyle{headings}
\mainmatter
\def\ECCVSubNumber{840}  

\title{DeepSFM: Structure From Motion Via Deep Bundle Adjustment} 

\titlerunning{DeepSFM}
%
\author{Xingkui Wei\inst{1}\thanks{indicates equal contributions.}  \and
Yinda Zhang\inst{2}\printfnsymbol{1} \and
Zhuwen Li\inst{3}\printfnsymbol{1}\and
Yanwei Fu\inst{1} \thanks{indicates corresponding author.}\and
Xiangyang Xue\inst{1}}
\authorrunning{X. Wei et al.}
%
\institute{$^1$Fudan University \samelineand $^2$Google Research  \samelineand $^3$Nuro, Inc}
\newcommand{\samelineand}{\qquad}

\maketitle

\begin{abstract}
Structure from motion (SfM) is an essential computer vision problem which has not been well handled by deep learning. One of the promising trends is to apply explicit structural constraint, e.g. 3D cost volume, into the network. However, existing methods usually assume accurate camera poses either from GT or other methods, which is unrealistic in practice.
In this work, we design a physical driven architecture, namely DeepSFM, inspired by traditional Bundle Adjustment (BA), which consists of two cost volume based architectures for depth and pose estimation respectively, iteratively running to improve both. 
The explicit constraints on both depth (structure) and pose (motion), when combined with the learning components, bring the merit from both traditional BA and emerging deep learning technology. Extensive experiments on various datasets show that our model achieves the state-of-the-art performance on both depth and pose estimation with superior robustness against less number of inputs and the noise in initialization.

\end{abstract}

\section{Introduction}

SfM is a fundamental human vision functionality which recovers 3D structures from the projected retinal images of moving objects or scenes. 
It enables machines to sense and understand the 3D world and is critical in achieving real-world artificial intelligence. 
Over decades of researches, there has been a lot of great success on SfM; however, the performance is far from perfect.

Conventional SfM approaches \cite{agarwal2011building,wu2011multicore,engel2017direct,delaunoy2014photometric} heavily rely on Bundle-Adjustment (BA) \cite{triggs1999bundle,agarwal2010bundle}, in which 3D structures and camera motions of each view are jointly optimized via Levenberg-Marquardt (LM) algorithm \cite{nocedal2006numerical} according to the cross-view correspondence.
Though successful in certain scenarios, conventional SfM based approaches are fundamentally restricted by the coverage of the provided multiple views and the overlaps among them.
They also typically fail to reconstruct textureless or non-lambertian (e.g. reflective or transparent) surfaces due to the missing of correspondence across views.
As a result, selecting sufficiently good input views and the right scene requires excessive caution and is usually non-trivial to even experienced user. 

\begin{figure}[tb]
\centering \includegraphics[width=122mm]{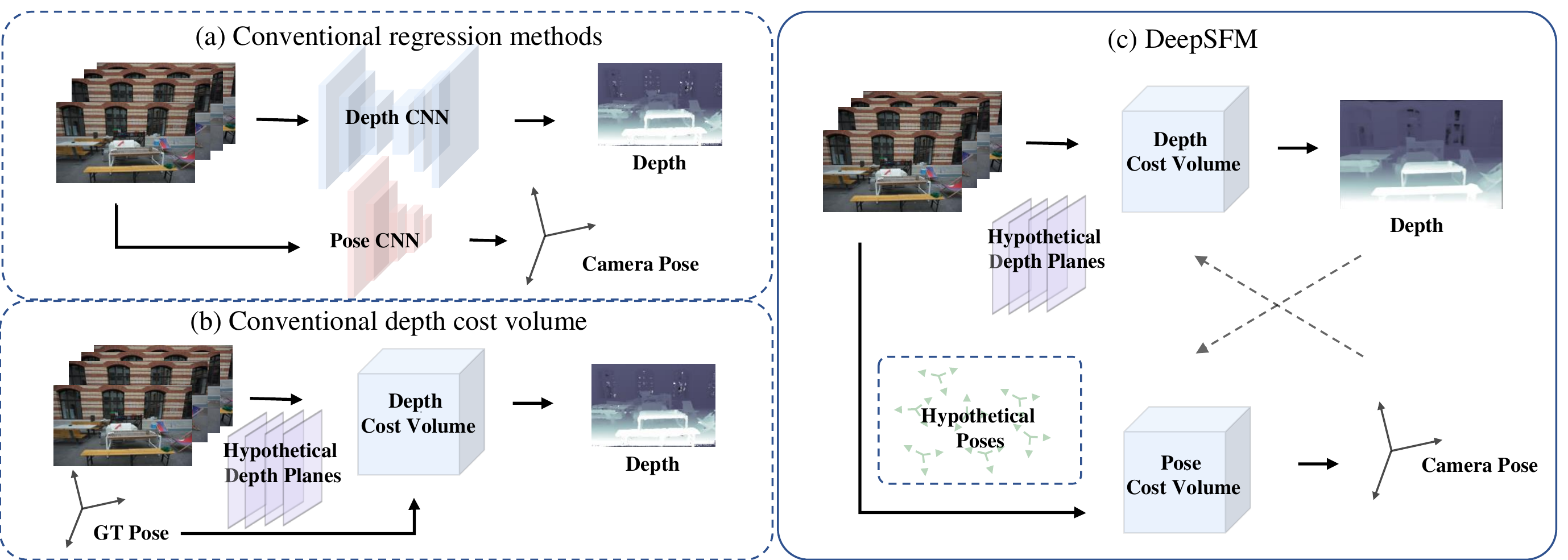}
\caption{DeepSFM refines the depth and camera pose of a target image given a few nearby source images. The network includes a depth based cost volume (D-CV) and a pose based cost volume (P-CV) which enforce photo-consistency and geometric-consistency into 3D cost volumes. The whole procedure is performed as iterations. }
\label{fig:intro} 
\end{figure}

Recent researches resort to deep learning to deal with the typical weakness of conventional SfM. 
Early effort utilizes deep neural network as a powerful mapping function that directly regresses the structures and motions \cite{ummenhofer2017demon,vijayanarasimhan2017sfm,zhou2017unsupervised,wang2017deepvo}. 
Since the geometric constraints of structures and motions are not explicitly enforced, the network does not learn the underlying physics and prone to overfitting.
Consequently, they do not perform as accurate as conventional SfM approaches and suffer from extremely poor generalization capability. Most recently, the 3D cost volume \cite{kar2017learning} has been introduced to explicit leveraging photo-consistency in a differentiable way, which significantly boosts the performance of deep learning based 3D reconstruction. 
However, the camera motion usually has to be known \cite{yao2018mvsnet,im2018dpsnet}, which requires to run traditional methods on densely captured high resolution images or relies on extra calibration devices (Fig. \ref{fig:intro} (b)). Some methods direct regress the motion \cite{ummenhofer2017demon,zhou2017unsupervised}, which still suffer from generalization issue (Fig. \ref{fig:intro} (a)).
Very rare deep learning approaches \cite{tang2018ba,teed2018deepv2d} can work well under noisy camera motion and improve both structure and motion simultaneously.

Inspired by BA and the success of cost volume for depth estimation, we propose a deep learning framework for SfM that iteratively improves both depth and camera pose according to cost volume explicitly built to measure photo-consistency and geometric-consistency.
Our method does not require accurate pose, and a rough estimation is enough.
In particular, our network includes a depth based cost volume (D-CV) and a pose based cost volume (P-CV). 
D-CV optimizes per-pixel depth values with the current camera poses, while P-CV optimizes camera poses with the current depth estimations (see Fig.\ref{fig:intro} (c)). 
Conventional 3D cost volume enforces photo-consistency by unprojecting pixels into the discrete camera fronto-parallel planes and computing the photometric (i.e. image feature) difference as the cost. In addition to that, our D-CV further enforces geometric-consistency among cameras with their current depth estimations by adding the geometric (i.e. depth) difference to the cost. Note that the initial depth estimation can be obtained using the conventional 3D cost volume. When preparing this work, we notice that a concurrent work \cite{xu2019multi} which also utilizes this trick to build a better cost volume in their system. 
For pose estimation, rather than direct regression, our P-CV discretizes around the current camera positions, and also computes the photometric and geometric differences by hypothetically moving the camera into the discretized position. Note that the initial camera pose can be obtained by a rough estimation from the direct regression methods such as \cite{ummenhofer2017demon}.
Our framework bridges the gap between the conventional and deep learning based SfM by incorporating explicit constraints of photo-consistency, geometric-consistency and camera motions all in the deep network.

The closest work in the literature is the recently proposed BA-Net \cite{tang2018ba}, which also aims to explicitly incorporate multi-view geometric constraints in a deep learning framework. They achieve this goal by integrating the LM optimization into the network. 
However, the LM iterations are unrolled with few iterations due to the memory and computational inefficiency, and thus it can potentially lead to non-optimal solutions due to lack of enough iterations. 
In contrast, our method does not have a restriction on the number of iterations and achieves empirically better performance.
Furthermore, LM in SfM originally optimizes point and camera positions, and thus direct integration of LM still requires good correspondences. To evade the correspondence issue in typical SfM, their models employ a direct regressor to predict depth at the front end, which heavily relies on prior in the training data. 
In contrast, our model is a fully physical-driven architecture that less suffers from over-fitting issue for both depth and pose estimation.

To demonstrate the superiority of our method, we conduct extensive experiments on \emph{DeMoN datasets}, \emph{ScanNet}, \emph{ETH3D} and \emph{Tanks and Temples}. The experiments show that our approach outperforms the state-of-the-art \cite{schonberger2016structure,ummenhofer2017demon,tang2018ba}.

\section{Related work}
\label{rel_work}

There is a large body of work that focuses on inferring depth or motion from color images, ranging from single view, multiple views and monocular video. We discuss them in the context of our work.

\textbf{Single-view Depth Estimation.}
While ill-posed, the emerging of deep learning technology enables the estimation of depth from a single color image.
The early work directly formulates this into a per-pixel regression problem \cite{eigen2014depth}, and follow-up works improve the performance by introducing multi-scale network architectures \cite{eigen2014depth,Eigen_2015_ICCV}, skip-connections \cite{Wang_2015_CVPR,liu2016learning}, powerful decoder and post process \cite{garg2016unsupervised,laina2016deeper,Kuznietsov_2017_CVPR,Wang_2015_CVPR,liu2016learning}, and new loss functions \cite{Fu_2018_CVPR}.
Even though single view based methods generate plausible results, the models usually resort heavily to the prior in the training data and suffer from generalization capability.
Nevertheless, these methods still act as an important component in some multi-view systems \cite{tang2018ba}. 


\textbf{Traditional Structure-from-Motion }
Simultaneously estimating 3d structure and camera motion is a well studied problem which has a traditional tool-chain of techniques \cite{furukawa2010towards,newcombe2011dtam,wu2011visualsfm}. Structure from Motion(SfM) has made great progress in many aspects. \cite{lowe2004distinctive,han2015matchnet} aim at improving features and \cite{snavely2011scene} introduce new optimization techniques. More robust structures and data representations are introduced by \cite{gherardi2010improving,schonberger2016structure}. Simultaneous Localization and Sapping(SLAM) systems track the motion of the camera and build 3D structure from video sequence \cite{newcombe2011dtam,engel2014lsd,mur2015orb,mur2017orb}. \cite{engel2014lsd} propose the photometric bundle adjustment algorithm to directly minimize the photometric error of aligned pixels. However, traditional SfM and SLAM methods are sensitive to low texture region, occlusions, moving objects and lighting changes, which limit the performance and stability.

\textbf{Deep Learning for Structure-from-Motion}
Deep neural networks have shown great success in stereo matching and Structure-from-Motion problems. \cite{ummenhofer2017demon,wang2017deepvo,vijayanarasimhan2017sfm,zhou2017unsupervised} regress depth and camera pose directly in a supervised manner or by introducing photometric constraints between depth and motion as a self-supervision signal. Such methods solve the camera motion as a regression problem, and the relation between camera motion and depth prediction is neglected.

Recently, some methods exploit multi-view photometric or feature-metric constraints to enforce the relationship between dense depth and the camera pose in network. The SE3 transformer layer is introduced by \cite{teed2018deepv2d}, which uses geometry to map flow and depth into a camera pose update. \cite{wang2018learning} propose the differentiable camera motion estimator based on the Direct Visual Odometry \cite{steinbrucker2011real}. \cite{clark2018learning} using a LSTM-RNN \cite{hochreiter2001learning} as the optimizer to solve nonlinear least squares in two-view SfM.  \cite{tang2018ba} train a network to generate a set of basis depth maps and optimize depth and camera poses in a BA-layer by minimizing a feature-metric error. 

\begin{figure*}[tb]
\centering \includegraphics[width=122mm]{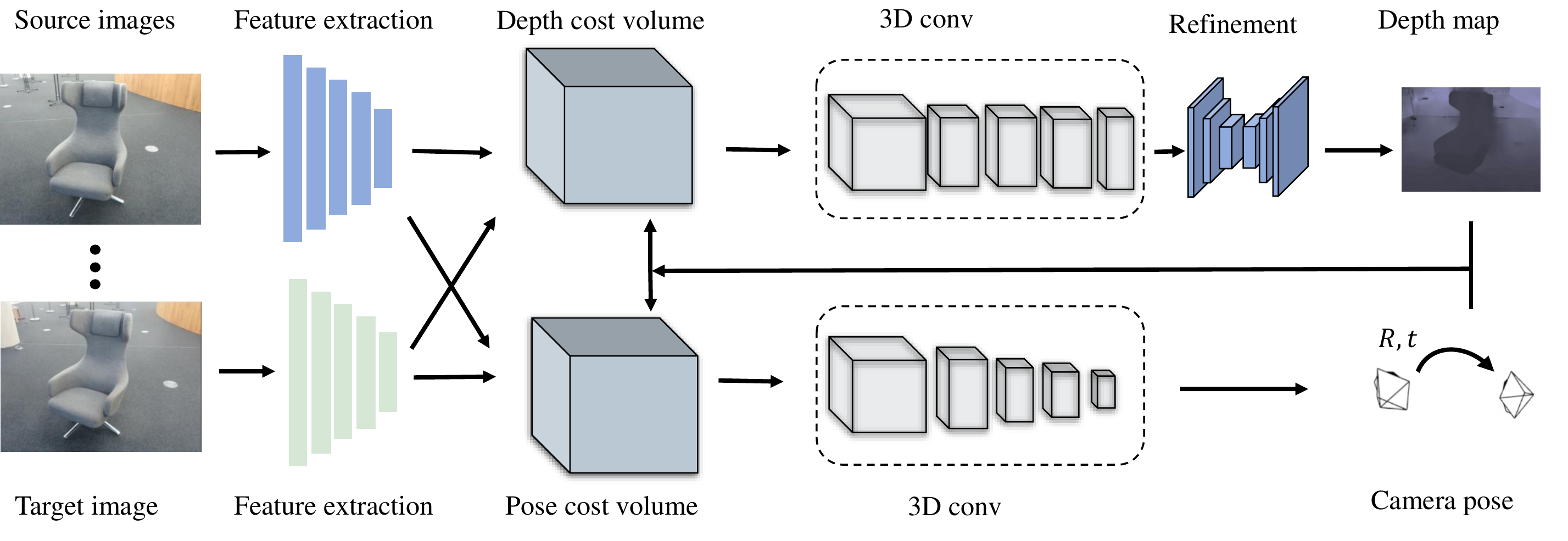}
\caption{Overview of DeepSFM. 2D CNN is used to extract photometric feature to construct cost volumes. Initial source depth maps and camera poses are used to introduce both photometric and geometric consistency. A series of 3D CNN layers are applied for D-CV and P-CV. Then a context network and depth regression operation are applied to produce predicted depth map of target image.}
\label{fig:pipeline} 
\end{figure*}

\section{Architecture}
\label{arch}
Our framework receives frames of a scene from different viewpoints, and produces 
accurate depth maps and camera poses for all frames. Similar to Bundle Adjustment (BA), we also assume initial structures (i.e depth maps) and motions (i.e. camera poses) are given. The initialization is not necessary to be accurate for the good performance using our framework and thus can be easily obtained from some direct regression based methods \cite{ummenhofer2017demon}. 

Now we introduce the overview of our model -- DeepSFM. 
Without loss of generality, we describe our model taking two images as inputs, namely the target image and the source image, 
and all the technical components can be extended for multiple images straightforwardly.
As shown in Fig.\ref{fig:pipeline}
, we first extract feature maps from inputs through a shared encoder. We then sample the solution space for depth uniformly in the inverse-depth space between a predefined range
, and camera pose around the initialization respectively. After that, we build cost volumes accordingly to reason the confidence of each depth and pose hypothesis.
This is achieved by validating the consistency between the feature of the target view and the ones warped from the source image.
Besides photo-metric consistency that measures the color image similarity, we also take into account the geometric consistency across warped depth maps.
Note that depth and pose require different designs of cost volume to efficiently sample the hypothesis space. {Gradients can back-propagate through cost volumes, and cost-volume construction does not affect any trainable parameters.}
The cost volumes are then fed into 3D CNN to regress new depth and pose.
These updated values can be used to create new cost volumes, and the model improves the prediction iteratively.


For notations, we denote $\{\textbf{I}_i\}_{i=1}^n$ as all the images in one scene, $\{\textbf{D}_i\}_{i=1}^n$  as the corresponding ground truth depth maps, $\{\textbf{K}_i\}_{i=1}^n$ as the camera intrinsics, $\{\textbf{R}_i,\textbf{t}_i\}_{i=1}^n$ as the ground truth rotations and translations of camera, $\{\textbf{D}^*_i\}_{i=1}^n$ and $\{\textbf{R}^*_i, \textbf{t}^*_i\}_{i=1}^n$ as initial depth maps and camera pose parameters for constructing cost volumes, where $n$ is the number of image samples. 

\subsection{2D Feature Extraction}
Given the input sequences $\{\textbf{I}_i\}_{i=1}^n$, we extract the 2D CNN feature $\{\textbf{F}_i\}_{i=1}^n$ for each frame. Firstly, a 7 layers' CNN with kernel size $3\times 3$ is applied to extract low contextual information. Then we adopt a spatial pyramid pooling (SPP) \cite{kaiming14ECCV} module, which can extract hierarchical multi-scale features through 4 average pooling blocks with different pooling kernel size ($4\times 4,8\times 8,16\times 16,32\times 32$). Finally, we pass the concatenated features through 2D CNNs to get the 32-channel image features after upsampling these multi-scale features into the same resolution. These image sequence features are used by the building of both our depth based and pose based cost volumes.

\subsection{Depth based Cost Volume (D-CV)}
Traditional plane sweep cost volume aims to back-project the source images onto successive virtual planes in the 3D space and measure photo-consistency error among the warped image features and target image features for each pixel. Different from the cost volume used in mainstream multi-view and structure-from-motion methods, we construct a D-CV to further utilize the local geometric consistency constraints introduced by depth maps. Inspired by the traditional plane sweep cost volumes, our D-CV is a concatenation of three components: the target image features, the warped source image features and the homogeneous depth consistency maps. 

\textbf{Hypothesis Sampling}
To back-project the features and depth maps from source viewpoint to the 3D space in target viewpoint, we uniformly sample a set of $L$ virtual planes $\{d_{l}\}_{l=1}^L$ in the inverse-depth space which are perpendicular to the forward direction ($z$-axis) of the target viewpoint. 
These planes serve as the hypothesis of the output depth map, and the cost volume can be built upon them.

\textbf{Feature warping} 
To construct our D-CV, we first warp source image features $\textbf{F}_i$ (of size $CHannel \times Width \times Height$ ) to each of the hypothetical depth map planes $d_{l}$ using camera intrinsic matrix $\textbf{K}$ and initial camera poses $\{\textbf{R}^*_i, \textbf{t}^*_i\}$, according to:
\begin{equation}
    \tilde{\mathbf{F}}_{i l}(u)=\mathbf{F}_{i}\left(\tilde{u}_{l}\right),
    \tilde{u}_{l} \sim \mathbf{K}\left[\mathbf{R}_{i}^{*} | \mathbf{t}_{i}^{*}\right] \left[ \begin{array}{c}{\left(\mathbf{K}^{-1} u\right) d_{l}} \\ {1}\end{array}\right]
   \label{eq:coord}
\end{equation}
where $u$ and $\tilde{u}_l$ are the homogeneous coordinates of each pixel in the target view and the projected coordinates onto the corresponding source view. $\tilde{\textbf{F}}_{i l}(u)$ denotes the warped feature of the source image through the $l$-th virtual depth plane. Note that the projected homogeneous coordinates $\tilde{u}_{l}$ are floating numbers, and we adopt a differentiable bilinear interpolation to generate the warped feature map $\tilde{\textbf{F}}_{il}$. The pixels with no source view coverage are assigned with zeros. Following \cite{im2018dpsnet}, we concatenate the target feature and the warped target feature together and obtain a $2CH \times L \times W \times H$ 4D feature volume.

\textbf{Depth consistency}
In addition to photometric consistency, to exploit geometric consistency and promote the quality of depth prediction, we add two more channels on each virtual plane: the warped initial depth maps from the source views and the projected virtual depth plane from the perspective of the source view. Note that the former is the same as image feature warping, while the latter requires a coordinate transformation from the target to the source camera. 

In particular, the first channel is computed as follows. The initial depth map of source image is first down-sampled and then warped to hypothetical depth planes similarly to the image feature warping as
$ \tilde{\textbf{D}}_{i l}^{*}(u)=\textbf{D}_{i}^{*}\left(\tilde{u}_{l}\right)$,
where the coordinates $u$ and $\tilde{u}_l$ are defined in Eq. \ref{eq:coord}  and $\tilde{\textbf{D}}_{i l}^{*}(u)$ represents the warped one-channel depth map on the $l$-th depth plane. { One distinction between depth warping and feature warping is that we adopt nearest neighbor sampling for depth warping, instead of bilinear interpolation. A comparison between the two methods is provided in the supplementary material. }

The second channel contains the depth values of the virtual planes in the target view by seeing them from the source view. 
To transform the virtual planes to the source view coordinate system, we apply a $T$ function on each virtual plane $d_l$ in the following:
\begin{equation}
    T(d_l) \sim \left[\mathbf{R}_{i}^{*} | \mathbf{t}_{i}^{*}\right] \left[ \begin{array}{c}{\left(\mathbf{K}^{-1} u\right) d_{l}} \\ {1}\end{array}\right]
\end{equation}
We stack the warped initial depth maps and the transformed depth planes together, and get a depth volume of size $2 \times L \times W \times H$. 

By concatenating the feature volume and depth volume together, we obtain a 4D cost tensor of size $\left(2CH + 2\right) \times L\times  W\times H$. Given the 4D cost volume, our network learns a cost volume of size $L\times W\times H$ using several 3D convolutional layers with kernel size $3\times3\times 3$. When there is more than one source image, we get the final cost volume by averaging over multiple input source views.

\begin{figure}[tb]\centering      
\subfigure[Camera translation sampling]{                                   \label{fig:trans_sampling}           
\includegraphics[scale=0.28]{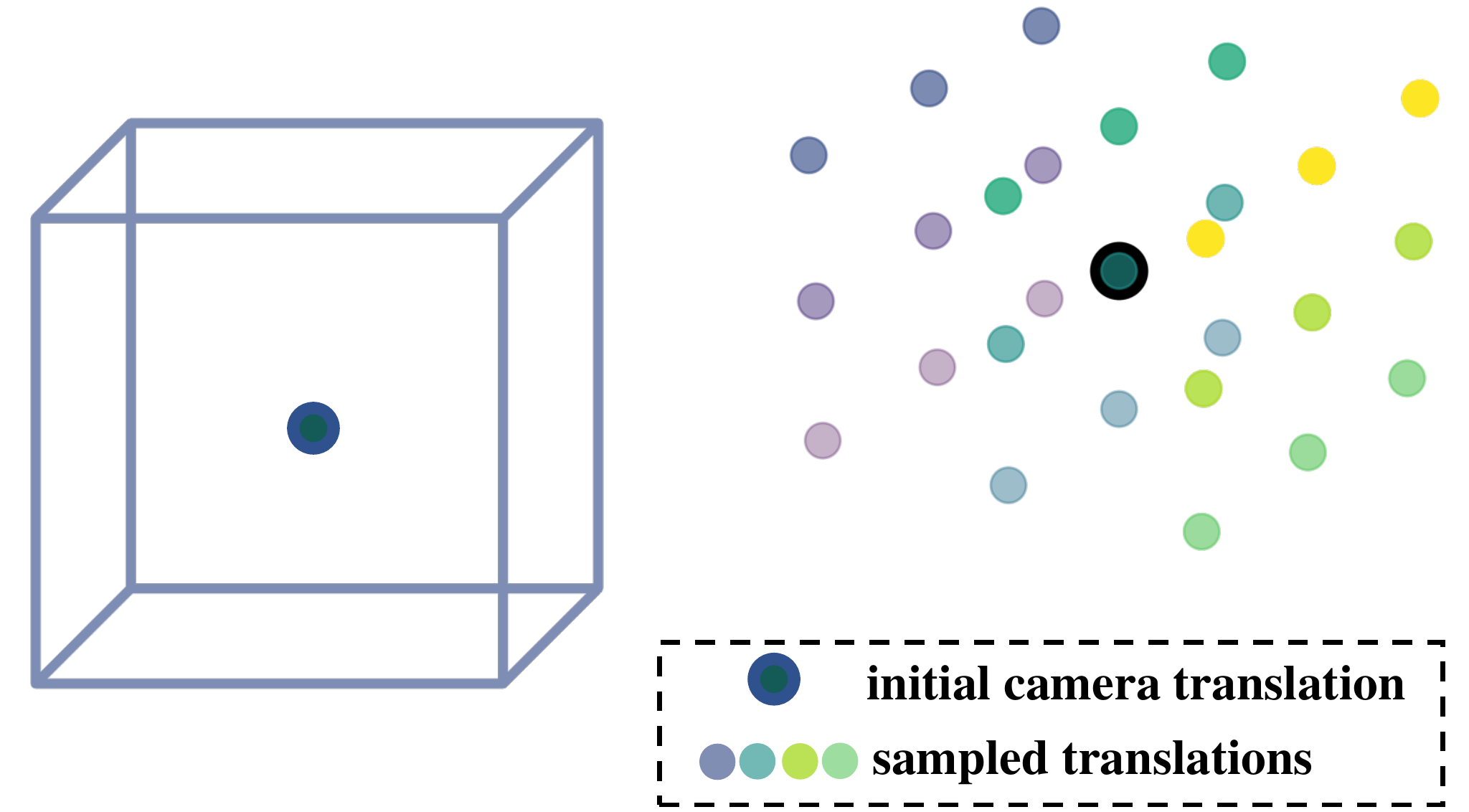}
}
\subfigure[Camera rotation sampling]{                    
    \label{fig:rot_sampling}            
\includegraphics[scale=0.28]{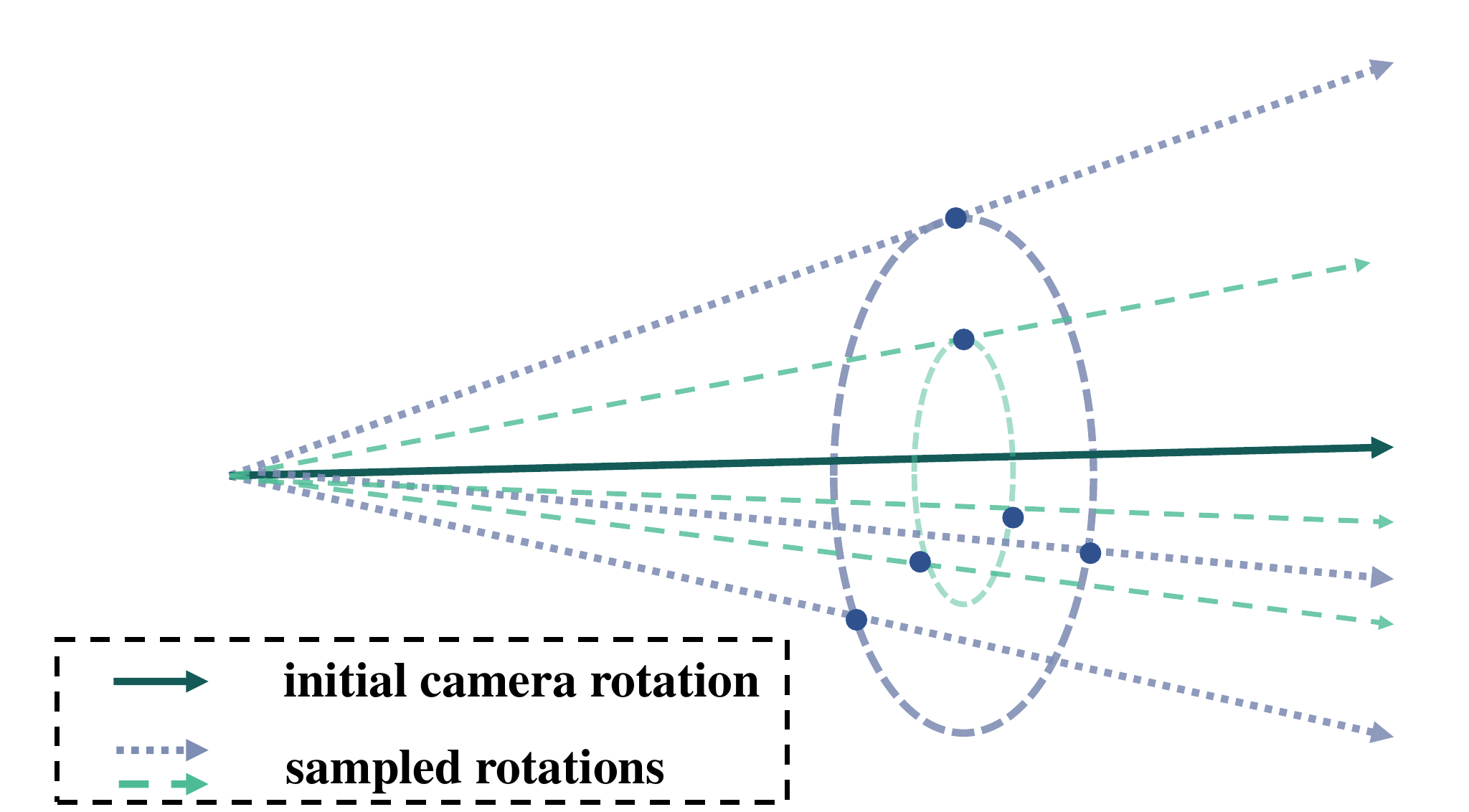}
}     
\caption{Hypothetical camera pose sampling. (a) Camera translation sampling. We sample uniformly in the cubic space. (b) Camera rotation sampling. We sample around initial orientation vector in conical space. }
\label{fig:p_cv} 
\end{figure}
\subsection{Pose based Cost Volume (P-CV)}
In addition to the construction of D-CV, we also propose a P-CV, aiming at optimizing initial camera poses through both photometric and geometric consistency (see Fig.\ref{fig:p_cv}). Instead of building a cost volume based on hypothetical depth map planes, our novel P-CV is constructed based on a set of assumptive camera poses. Similar to D-CV, P-CV is also concatenated by three components: the target image features, the warped source image features and the homogeneous depth consistency maps. Given initial camera pose parameters $\{\textbf{R}^*_i, \textbf{t}^*_i\}$, we uniformly sample a batch of discrete candidate camera poses around. { As shown in Fig.\ref{fig:p_cv}, we shift rotation and translation separately while keeping the other one unchanged. For rotation, we sample $\delta R$ uniformly in the Euler angle space in a predefined range and multiply $\delta R$ by the initial $R$. For translation, we sample $\delta t$ uniformly and add $\delta t$ to the initial $t$.
} In the end, a group of $P$ virtual camera poses noted as $\{\textbf{R}^*_{i p}|\textbf{t}^*_{i p}\}^P_{p=1}$ around input pose are obtained for cost volume construction.

The posed-based cost volume is also constructed by concatenating image features and homogeneous depth maps. However, source view features and depth maps are warped based on sampled camera poses. For feature warping, we compute $\tilde{u}_{p}$ as following equations: 
\begin{equation}
   \tilde{u}_p \sim \mathbf{K}\left[\textbf{R}^*_{i p}|\textbf{t}^*_{i p}\right] \left[ \begin{array}{c}{\left(\mathbf{K}^{-1} u\right)\textbf{D}^*_i} \\ {1}\end{array}\right]
   \label{eq:poseproj}
\end{equation}
where $\textbf{D}^*_i$ is the initial target view depth. Similar to D-CV, we get warped source feature map $\tilde{\textbf{F}}_{ip}$ after bilinear sampling and concatenate it with target view feature map. We also transform the initial target view depth and source view depth into one homogeneous coordinate system, which enhances the geometric consistency between camera pose and multi view depth maps. 

After concatenating the above feature maps and depth maps together, we again build a 4D cost volume of size $\left(2CH+ 2\right) \times P \times W\times H$, where $W$ and $H$ are the width and height of feature map, $CH$ is the number of channels. We get output of size $1 \times P \times 1\times 1$ from the above 4-D tensor after eight 3D convolutional layers with kernel size $3\times3\times 3$, three 3D average pooling layers with stride size $2 \times 2 \times 1$ and one global average pooling at the end. 

\subsection{Cost Aggregation and Regression}
For depth prediction, we follow the cost aggregation technique introduced by \cite{im2018dpsnet}. We adopt a context network, which takes target image features and each slice of the coarse cost volume after 3D convolution as input and produce the refined cost slice. The final aggregated depth based volume is obtained by adding coarse and refined cost slices together. 
The last step to get depth prediction of target image is depth regression by soft-argmax as proposed in \cite{im2018dpsnet}. 
For camera poses prediction, we also apply a soft-argmax function on pose cost volume and get the estimated output rotation and translation vectors. 

\subsection{Training}
The DeepSFM learns the feature extractor, 3D convolution, and the regression layers in a supervised way. We denote $\hat{\textbf{R}}_i$ and $\hat{\textbf{t}}_i$ as predicted rotation angles and translation vectors of camera pose. Then the pose loss $\mathcal{L}_{rotation}$ is defined as the $L1$ distance between prediction and groundtruth. We denote $\hat{D}_i^0$ and $\hat{D}_i$ as predicted coarse depth map and refined depth map, then the depth loss function is defined as $
    \mathcal{L}_{depth}=\sum_i \lambda H(\hat{D}_i^0, \textbf{D}_i)+H(\hat{D}_i, \textbf{D}_i)$, where $\lambda$ is weight parameter and function $H$ is Huber loss. Our final objective $\mathcal{L}_{final}=\lambda_r\mathcal{L}_{rotation}+\lambda_t\mathcal{L}_{translation}+\lambda_d\mathcal{L}_{depth}
$. {The $\lambda$s are determined empirically,
and are listed in the supplementary material.}

The initial depth maps and camera poses are obtained from DeMoN. To keep correct scale, we multiply translation vectors and depth maps by the norm of the ground truth camera translation. The whole training and testing procedure are performed as four iterations. During each iteration, we take the predicted depth maps and camera poses of previous iteration as new initialization. {More details are provided in the supplementary material.}

\section{Experiments}
\label{eva}
\subsection{Datasets}
We evaluate DeepSFM on widely used datasets and compare with state-of-the-art methods on accuracy, generalization capability and robustness to initialization.

\textbf{DeMoN Datasets} \cite{ummenhofer2017demon} This dataset contains data from various sources, including SUN3D \cite{xiao2013sun3d}, RGB-D SLAM \cite{sturm2012benchmark}, and Scenes11 \cite{chang2015shapenet}. 
To test the generalization capability, we also evaluate on MVS \cite{fuhrmann2014mve} dataset but not use it for the training.
In all four datasets, RGB image sequences and the ground truth depth maps are provided with the camera intrinsics and camera poses.
Note that those datasets together provide a diverse set of both indoor and outdoor, synthetic and real-world scenes. 
{For all the experiments, we adopt the same training and testing data split from DeMoN.} 

\textbf{ETH3D Dataset } \cite{schoeps2017cvpr} It provides a variety of indoor and outdoor scenes with high-precision ground truth 3D points captured by laser scanners, which is a more solid benchmark dataset. 
Ground truth depth maps are obtained by projecting the point clouds to each camera view. Raw images are in high resolution but resized to $810\times540$ pixels for evaluation \cite{im2018dpsnet}. 

\textbf{Tanks and Temples} 
\cite{Knapitsch2017} It is a benchmark for image-based large scale 3D reconstruction. The benchmark sequences are acquired in realistic conditions and of high quality. Point clouds captured using an industrial laser scanner are provided as ground truth. Again, our method are trained on DeMoN and tested on the dataset to show the robustness to noisy initialization.

\begin{figure*}[tb]
\centering \includegraphics[width=122mm]{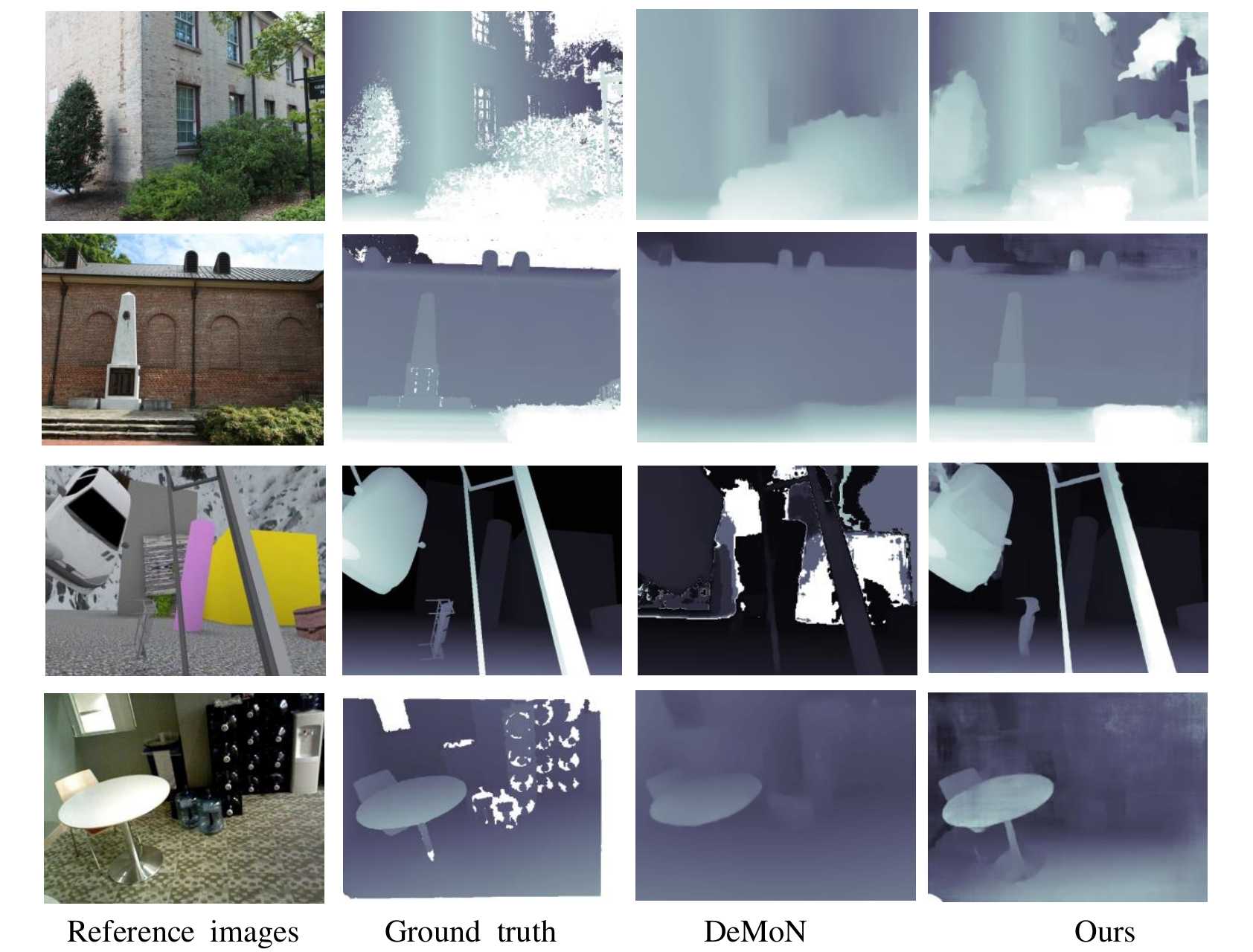}
\caption{Qualitative Comparisons with DeMoN \cite{ummenhofer2017demon} on DeMoN datasets. {Results on more methods and examples are shown in the supplementary material.}}
\label{fig:vis_demon} 
\end{figure*}
\begin{table*}[tb]
\begin{centering}
\caption{\label{tab:demon}Results on DeMoN datasets, the best results are noted by \textbf{Bold}.}
\begin{tabular}{cr@{\extracolsep{0pt}.}lr@{\extracolsep{0pt}.}lr@{\extracolsep{0pt}.}lr@{\extracolsep{0pt}.}lr@{\extracolsep{0pt}.}lr@{\extracolsep{0pt}.}lr@{\extracolsep{0pt}.}lr@{\extracolsep{0pt}.}lr@{\extracolsep{0pt}.}lr@{\extracolsep{0pt}.}lr@{\extracolsep{0pt}.}l}
\toprule[2pt] 
\textbf{\scriptsize{}MVS} & \multicolumn{6}{c}{{\scriptsize{}Depth}} & \multicolumn{4}{c}{{\scriptsize{}Motion}} & \multicolumn{2}{c}{\textbf{\scriptsize{}Scenes11}} & \multicolumn{6}{c}{{\scriptsize{}Depth}} & \multicolumn{4}{c}{{\scriptsize{}Motion}}\tabularnewline
\midrule
{\scriptsize{}Method} & \multicolumn{2}{c}{{\scriptsize{}L1-inv}} & \multicolumn{2}{c}{{\scriptsize{}sc-inv}} & \multicolumn{2}{c}{{\scriptsize{}L1-rel}} & \multicolumn{2}{c}{{\scriptsize{}Rot}} & \multicolumn{2}{c}{{\scriptsize{}Trans}} & \multicolumn{2}{c}{{\scriptsize{}Method}} & \multicolumn{2}{c}{{\scriptsize{}L1-inv}} & \multicolumn{2}{c}{{\scriptsize{}sc-inv}} & \multicolumn{2}{c}{{\scriptsize{}L1-rel}} & \multicolumn{2}{c}{{\scriptsize{}Rot}} & \multicolumn{2}{c}{{\scriptsize{}Trans}}\tabularnewline
\midrule[2pt]  
{\scriptsize{}Base-Oracle} & {\scriptsize{}0}&{\scriptsize{}019} & {\scriptsize{}0}&{\scriptsize{}197} & {\scriptsize{}0}&{\scriptsize{}105} & \multicolumn{2}{c}{{\scriptsize{}0}} & \multicolumn{2}{c}{{\scriptsize{}0}} & \multicolumn{2}{c}{{\scriptsize{}Base-Oracle}} & {\scriptsize{}0}&{\scriptsize{}023} & {\scriptsize{}0}&{\scriptsize{}618} & {\scriptsize{}0}&{\scriptsize{}349} & \multicolumn{2}{c}{{\scriptsize{}0}} & \multicolumn{2}{c}{{\scriptsize{}0}}\tabularnewline
{\scriptsize{}Base-SIFT} & {\scriptsize{}0}&{\scriptsize{}056} & {\scriptsize{}0}&{\scriptsize{}309} & {\scriptsize{}0}&{\scriptsize{}361} & {\scriptsize{}21}&{\scriptsize{}180} & {\scriptsize{}60}&{\scriptsize{}516} & \multicolumn{2}{c}{{\scriptsize{}Base-SIFT}} & {\scriptsize{}0}&{\scriptsize{}051} & {\scriptsize{}0}&{\scriptsize{}900} & {\scriptsize{}1}&{\scriptsize{}027} & {\scriptsize{}6}&{\scriptsize{}179} & {\scriptsize{}56}&{\scriptsize{}650}\tabularnewline
{\scriptsize{}Base-FF} & {\scriptsize{}0}&{\scriptsize{}055} & {\scriptsize{}0}&{\scriptsize{}308} & {\scriptsize{}0}&{\scriptsize{}322} & {\scriptsize{}4}&{\scriptsize{}834} & {\scriptsize{}17}&{\scriptsize{}252} & \multicolumn{2}{c}{{\scriptsize{}Base-FF}} & {\scriptsize{}0}&{\scriptsize{}038} & {\scriptsize{}0}&{\scriptsize{}793} & {\scriptsize{}0}&{\scriptsize{}776} & {\scriptsize{}1}&{\scriptsize{}309} & {\scriptsize{}19}&{\scriptsize{}426}\tabularnewline
{\scriptsize{}Base-Matlab} & \multicolumn{2}{c}{{\scriptsize{}-}} & \multicolumn{2}{c}{{\scriptsize{}-}} & \multicolumn{2}{c}{{\scriptsize{}-}} & {\scriptsize{}10}&{\scriptsize{}843} & {\scriptsize{}32}&{\scriptsize{}736} & \multicolumn{2}{c}{{\scriptsize{}Base-Matlab}} & \multicolumn{2}{c}{{\scriptsize{}-}} & \multicolumn{2}{c}{{\scriptsize{}-}} & \multicolumn{2}{c}{{\scriptsize{}-}} & {\scriptsize{}0}&{\scriptsize{}917} & {\scriptsize{}14}&{\scriptsize{}639}\tabularnewline
{\scriptsize{}DeMoN} & {\scriptsize{}0}&{\scriptsize{}047} & {\scriptsize{}0}&{\scriptsize{}202} & {\scriptsize{}0}&{\scriptsize{}305} & {\scriptsize{}5}&{\scriptsize{}156} & {\scriptsize{}14}&{\scriptsize{}447} & \multicolumn{2}{c}{{\scriptsize{}DeMoN}} & {\scriptsize{}0}&{\scriptsize{}019} & {\scriptsize{}0}&{\scriptsize{}315} & {\scriptsize{}0}&{\scriptsize{}248} & {\scriptsize{}0}&{\scriptsize{}809} & {\scriptsize{}8}&{\scriptsize{}918}\tabularnewline
{\scriptsize{}LS-Net} & {\scriptsize{}0}&{\scriptsize{}051} & {\scriptsize{}0}&{\scriptsize{}221} & {\scriptsize{}0}&{\scriptsize{}311} & {\scriptsize{}4}&{\scriptsize{}653} & {\scriptsize{}11}&{\scriptsize{}221} & \multicolumn{2}{c}{{\scriptsize{}LS-Net}} & {\scriptsize{}0}&{\scriptsize{}010} & {\scriptsize{}0}&{\scriptsize{}410} & {\scriptsize{}0}&{\scriptsize{}210} & {\scriptsize{}4}&{\scriptsize{}653} & {\scriptsize{}8}&{\scriptsize{}210}\tabularnewline
{\scriptsize{}BANet} & {\scriptsize{}0}&{\scriptsize{}030} & {\scriptsize{}0}&{\scriptsize{}150} & {\scriptsize{}0}&{\scriptsize{}080} & {\scriptsize{}3}&{\scriptsize{}499} & {\scriptsize{}11}&{\scriptsize{}238} & \multicolumn{2}{c}{{\scriptsize{}BANet}} & {\scriptsize{}0}&{\scriptsize{}080} & {\scriptsize{}0}&{\scriptsize{}210} & {\scriptsize{}0}&{\scriptsize{}130} & {\scriptsize{}3}&{\scriptsize{}499} & {\scriptsize{}10}&{\scriptsize{}370}\tabularnewline
{\scriptsize{}Ours} & \textbf{\scriptsize{}0}&\textbf{\scriptsize{}021} & \textbf{\scriptsize{}0}&\textbf{\scriptsize{}129} & \textbf{\scriptsize{}0}&\textbf{\scriptsize{}079} & \textbf{\scriptsize{}2}&\textbf{\scriptsize{}824} & \textbf{\scriptsize{}9}&\textbf{\scriptsize{}881} & \multicolumn{2}{c}{{\scriptsize{}Ours}} & \textbf{\scriptsize{}0}&\textbf{\scriptsize{}007} & \textbf{\scriptsize{}0}&\textbf{\scriptsize{}112} & \textbf{\scriptsize{}0}&\textbf{\scriptsize{}064} & \textbf{\scriptsize{}0}&\textbf{\scriptsize{}403} & \textbf{\scriptsize{}5}&\textbf{\scriptsize{}828}\tabularnewline
\midrule[2pt]  
\textbf{\scriptsize{}RGB-D} & \multicolumn{6}{c}{{\scriptsize{}Depth}} & \multicolumn{4}{c}{{\scriptsize{}Motion}} & \multicolumn{2}{c}{\textbf{\scriptsize{}Sun3D}} & \multicolumn{6}{c}{{\scriptsize{}Depth}} & \multicolumn{4}{c}{{\scriptsize{}Motion}}\tabularnewline
\midrule 
{\scriptsize{}Method} & \multicolumn{2}{c}{{\scriptsize{}L1-inv}} & \multicolumn{2}{c}{{\scriptsize{}sc-inv}} & \multicolumn{2}{c}{{\scriptsize{}L1-rel}} & \multicolumn{2}{c}{{\scriptsize{}Rot}} & \multicolumn{2}{c}{{\scriptsize{}Trans}} & \multicolumn{2}{c}{{\scriptsize{}Method}} & \multicolumn{2}{c}{{\scriptsize{}L1-inv}} & \multicolumn{2}{c}{{\scriptsize{}sc-inv}} & \multicolumn{2}{c}{{\scriptsize{}L1-rel}} & \multicolumn{2}{c}{{\scriptsize{}Rot}} & \multicolumn{2}{c}{{\scriptsize{}Trans}}\tabularnewline
\midrule[2pt] 
{\scriptsize{}Base-Oracle} & {\scriptsize{}0}&{\scriptsize{}026} & {\scriptsize{}0}&{\scriptsize{}398} & {\scriptsize{}0}&{\scriptsize{}36} & \multicolumn{2}{c}{{\scriptsize{}0}} & \multicolumn{2}{c}{{\scriptsize{}0}} & \multicolumn{2}{c}{{\scriptsize{}Base-Oracle}} & {\scriptsize{}0}&{\scriptsize{}020} & {\scriptsize{}0}&{\scriptsize{}241} & {\scriptsize{}0}&{\scriptsize{}220} & \multicolumn{2}{c}{{\scriptsize{}0}} & \multicolumn{2}{c}{{\scriptsize{}0}}\tabularnewline
{\scriptsize{}Base-SIFT} & {\scriptsize{}0}&{\scriptsize{}050} & {\scriptsize{}0}&{\scriptsize{}577} & {\scriptsize{}0}&{\scriptsize{}703} & {\scriptsize{}12}&{\scriptsize{}010} & {\scriptsize{}56}&{\scriptsize{}021} & \multicolumn{2}{c}{{\scriptsize{}Base-SIFT}} & {\scriptsize{}0}&{\scriptsize{}029} & {\scriptsize{}0}&{\scriptsize{}290} & {\scriptsize{}0}&{\scriptsize{}286} & {\scriptsize{}7}&{\scriptsize{}702} & {\scriptsize{}41}&{\scriptsize{}825}\tabularnewline
{\scriptsize{}Base-FF} & {\scriptsize{}0}&{\scriptsize{}045} & {\scriptsize{}0}&{\scriptsize{}548} & {\scriptsize{}0}&{\scriptsize{}613} & {\scriptsize{}4}&{\scriptsize{}709} & {\scriptsize{}46}&{\scriptsize{}058} & \multicolumn{2}{c}{{\scriptsize{}Base-FF}} & {\scriptsize{}0}&{\scriptsize{}029} & {\scriptsize{}0}&{\scriptsize{}284} & {\scriptsize{}0}&{\scriptsize{}297} & {\scriptsize{}3}&{\scriptsize{}681} & {\scriptsize{}33}&{\scriptsize{}301}\tabularnewline
{\scriptsize{}Base-Matlab} & \multicolumn{2}{c}{{\scriptsize{}-}} & \multicolumn{2}{c}{{\scriptsize{}-}} & \multicolumn{2}{c}{{\scriptsize{}-}} & {\scriptsize{}12}&{\scriptsize{}813} & {\scriptsize{}49}&{\scriptsize{}612} & \multicolumn{2}{c}{{\scriptsize{}Base-Matlab}} & \multicolumn{2}{c}{{\scriptsize{}-}} & \multicolumn{2}{c}{{\scriptsize{}-}} & \multicolumn{2}{c}{{\scriptsize{}-}} & {\scriptsize{}5}&{\scriptsize{}920} & {\scriptsize{}32}&{\scriptsize{}298}\tabularnewline
{\scriptsize{}DeMoN} & {\scriptsize{}0}&{\scriptsize{}028} & {\scriptsize{}0}&{\scriptsize{}130} & {\scriptsize{}0}&{\scriptsize{}212} & {\scriptsize{}2}&{\scriptsize{}641} & {\scriptsize{}20}&{\scriptsize{}585} & \multicolumn{2}{c}{{\scriptsize{}DeMoN}} & {\scriptsize{}0}&{\scriptsize{}019} & {\scriptsize{}0}&{\scriptsize{}114} & {\scriptsize{}0}&{\scriptsize{}172} & {\scriptsize{}1}&{\scriptsize{}801} & {\scriptsize{}18}&{\scriptsize{}811}\tabularnewline
{\scriptsize{}LS-Net} & {\scriptsize{}0}&{\scriptsize{}019} & {\scriptsize{}0}&{\scriptsize{}090} & {\scriptsize{}0}&{\scriptsize{}301} & \textbf{\scriptsize{}1}&\textbf{\scriptsize{}010} & {\scriptsize{}22}&{\scriptsize{}100} & \multicolumn{2}{c}{{\scriptsize{}LS-Net}} & {\scriptsize{}0}&{\scriptsize{}015} & {\scriptsize{}0}&{\scriptsize{}189} & {\scriptsize{}0}&{\scriptsize{}650} & \textbf{\scriptsize{}1}&\textbf{\scriptsize{}521} & {\scriptsize{}14}&{\scriptsize{}347}\tabularnewline
{\scriptsize{}BANet} & \textbf{\scriptsize{}0}&\textbf{\scriptsize{}008} & {\scriptsize{}0}&{\scriptsize{}087} & \textbf{\scriptsize{}0}&\textbf{\scriptsize{}050} & {\scriptsize{}2}&{\scriptsize{}459} & {\scriptsize{}14}&{\scriptsize{}900} & \multicolumn{2}{c}{{\scriptsize{}BANet}} & {\scriptsize{}0}&{\scriptsize{}015} & {\scriptsize{}0}&{\scriptsize{}110} & \textbf{\scriptsize{}0}&\textbf{\scriptsize{}060} & {\scriptsize{}1}&{\scriptsize{}729} & {\scriptsize{}13}&{\scriptsize{}260}\tabularnewline
{\scriptsize{}Ours} & {\scriptsize{}0}&{\scriptsize{}011} & \textbf{\scriptsize{}0}&\textbf{\scriptsize{}071} & {\scriptsize{}0}&{\scriptsize{}126} & {\scriptsize{}1}&{\scriptsize{}862} & \textbf{\scriptsize{}14}&\textbf{\scriptsize{}570} & \multicolumn{2}{c}{{\scriptsize{}Ours}} & \textbf{\scriptsize{}0}&\textbf{\scriptsize{}013} & \textbf{\scriptsize{}0}&\textbf{\scriptsize{}093} & {\scriptsize{}0}&{\scriptsize{}072} & {\scriptsize{}1}&{\scriptsize{}704} & \textbf{\scriptsize{}13}&\textbf{\scriptsize{}107}\tabularnewline
\bottomrule[2pt] 
\end{tabular}

\end{centering}

\end{table*}

\begin{figure*}[tb]
\centering \includegraphics[width=122mm]{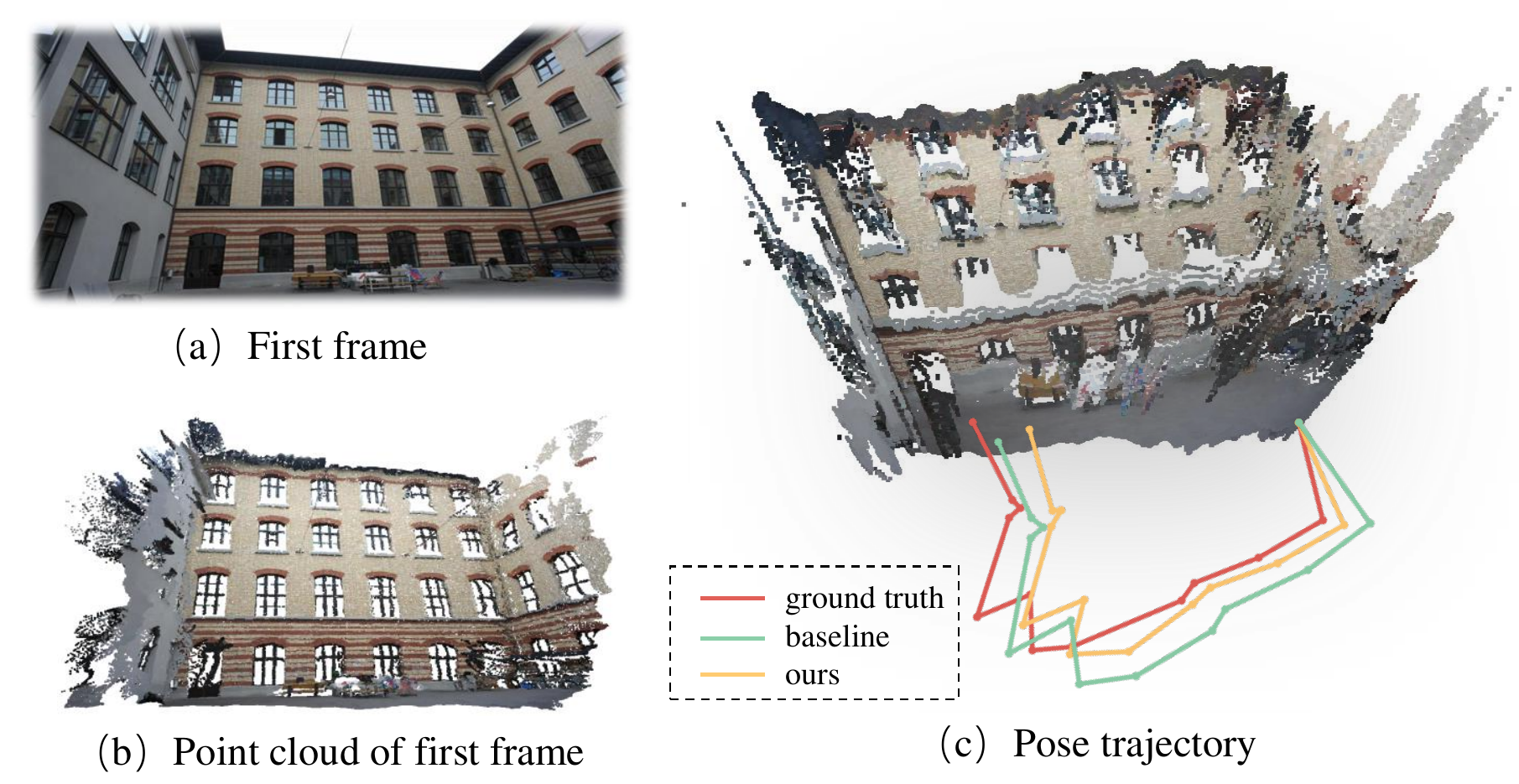}
\caption{Result on a sequence of the ETH3D dataset. (a) First frame of sequence. (b) The point cloud from estimated depth of the first frame. (c) Pose trajectories. Compared with the baseline method in sec.\ref{sec:effect_pcv}, the accumulated pairwise pose trajectory predicted by our network (yellow) are more closely consistent with the ground truth (red). }
\label{fig:eth3d_pose} 
\end{figure*}
\begin{table*}[tb]
\begin{center}
\caption{\label{tab:ETH3D}Results on ETH3D (\textbf{Bold}: best; $\alpha=1.25$). abs\_rel, abs\_diff, sq\_rel, rms, and log\_rms, are absolute relative error, absolute difference, square relative difference, root mean square and log root mean square, respectively. }
\begin{tabular}{ccccccccc}
\toprule[2px] 
\multirow{2}{*}{Method} & \multicolumn{5}{c}{Error metric} & \multicolumn{3}{c}{Accuracy metric($\delta<\alpha^t$)}\tabularnewline
\cmidrule{2-9} \cmidrule{3-9} \cmidrule{4-9} \cmidrule{5-9} \cmidrule{6-9} \cmidrule{7-9} \cmidrule{8-9} \cmidrule{9-9} 
 & abs\_rel & abs\_diff & sq\_rel & rms & log\_rms & $\alpha$ & $\alpha^2$ & $\alpha^3$\tabularnewline
\midrule[2px]
COLMAP & 0.324 & \textbf{0.615} & 36.71 & 2.370 & 0.349 & \textbf{86.5} & 90.3 & 92.7\tabularnewline
DeMoN & 0.191 & 0.726 & 0.365 & 1.059 & 0.240 & 73.3 & 89.8 & 95.1\tabularnewline
Ours & \textbf{0.127} & 0.661 & \textbf{0.278} & \textbf{1.003} & \textbf{0.195} & 84.1 & \textbf{93.8} & \textbf{96.9}\tabularnewline
\bottomrule[2px] 
\end{tabular}
\end{center}
\end{table*}

\subsection{Evaluation}
\textbf{DeMoN Datasets} Our results on DeMoN datasets and the comparison to other methods are shown in Table \ref{tab:demon}. We cite results of some strong baseline methods from DeMoN paper, named as Base-Oracle, Base-SIFT, Base-FF and Base-Matlab respectively \cite{ummenhofer2017demon}. Base-Oracle estimate depth with the ground truth camera motion using SGM \cite{hirschmuller2005accurate}.
Base-SIFT, Base-FF and Base-Matlab solve camera motion and depth using feature, optical flow, and KLT tracking correspondence from 8-pt algorithm \cite{hartley1997defense}. We also compare to some most recent state-of-the-art methods LS-Net \cite{clark2018learning} and BA-Net \cite{tang2018ba}. {LS-Net introduces the learned LSTM-RNN optimizer to minimizing photometric error for stereo reconstruction. BA-Net is the most recent work that minimizes the feature-metric error between multi-view via the differentiable Levenberg-Marquardt \cite{lourakis2005levenberg} algorithm.} To make a fair comparison, we adopt the same metrics as DeMoN\cite{ummenhofer2017demon} for evaluation. 

Our method outperforms all traditional baseline methods and DeMoN on both depth and camera poses.
When compared with more recent LS-Net and BA-Net, our method produces better results in most metrics on four datasets.
On RGB-D dataset, our performance is comparable to the state-of-the-art due to relatively higher noise in the RGB-D ground truth. {LS-Net trains an initialization network which regresses depth and motion directly before adding the LSTM-RNN optimizer. The performance of the RNN optimizer is highly affected by the accuracy of the regressed initialization. The depth results of LS-Net are consistently poorer than BA-Net and our method, despite better rotation parameters are estimated by LS-Net on RGB-D and Sun3D datasets with very good initialization. Our method is slightly inferior to BA-Net on the L1-rel metric, which is probably due to that we sample 64 virtual planes uniformly as the hypothetical depth set, while BA-Net optimizes depth prediction based on a set of 128-channel estimated basis depth maps that are more memory consuming but have more fine-grained results empirically.}
Despite all that, it is shown that our learned cost volumes with geometric consistency work better than the photometric bundle adjustment (e.g. used in BA-Net) in most scenes.
In particular, we improve mostly on the Scenes11 dataset, where the ground truth is perfect but the input images contain a lot of texture-less regions, which are challenging to photo-consistency based methods. The Qualitative Comparisons between our method and DeMoN are shown in Fig.\ref{fig:vis_demon}.

\textbf{ETH3D} We further test the generalization capability on ETH3D. We provide comparisons to COLMAP \cite{schonberger2016structure} and DeMoN on ETH3D. COLMAP is a state-of-the-art Structure-from-Motion method, while DeMoN introduces a classical deep network architecture that directly regress depth and motion in a supervised manner. {Note that all the models are trained on DeMoN and then tested on the data provided by \cite{huang2018deepmvs}.} In the accuracy metric, the error $\delta$ s defined as $\textrm{max}(\frac{y_i^*}{y_i}, \frac{y_i}{y_i^*})$, and the thresholds are typically set as $[1.25,1.25^2,1.25^3]$. In Table \ref{tab:ETH3D}, our method shows the best performance overall among all the comparison methods. Our method produces better results than DeMoN consistently, since we impose geometric and physical constraints onto network rather than learning to regress directly. When compared with COLMAP, our method performs better on most metrics. COLMAP behaves well in the accuracy metric (i.e. abs\_diff). However, the presence of outliers is often observed in the predictions of COLMAP, which leads to poor performance in other metrics such as abs\_rel and sq\_rel, since those metrics are sensitive to outliers. As an intuitive display, we compute the motion of camera in a selected image sequence of ETH3D, as shown in Fig. \ref{fig:eth3d_pose}c. The point cloud computed from the estimated depth map is showed in Fig.\ref{fig:eth3d_pose}b, which is of good quality. 

\begin{figure}[tb]\centering 

\subfigure[resolution downscale]{       \centering    
\label{fig:resd}           
\includegraphics[width=54mm]{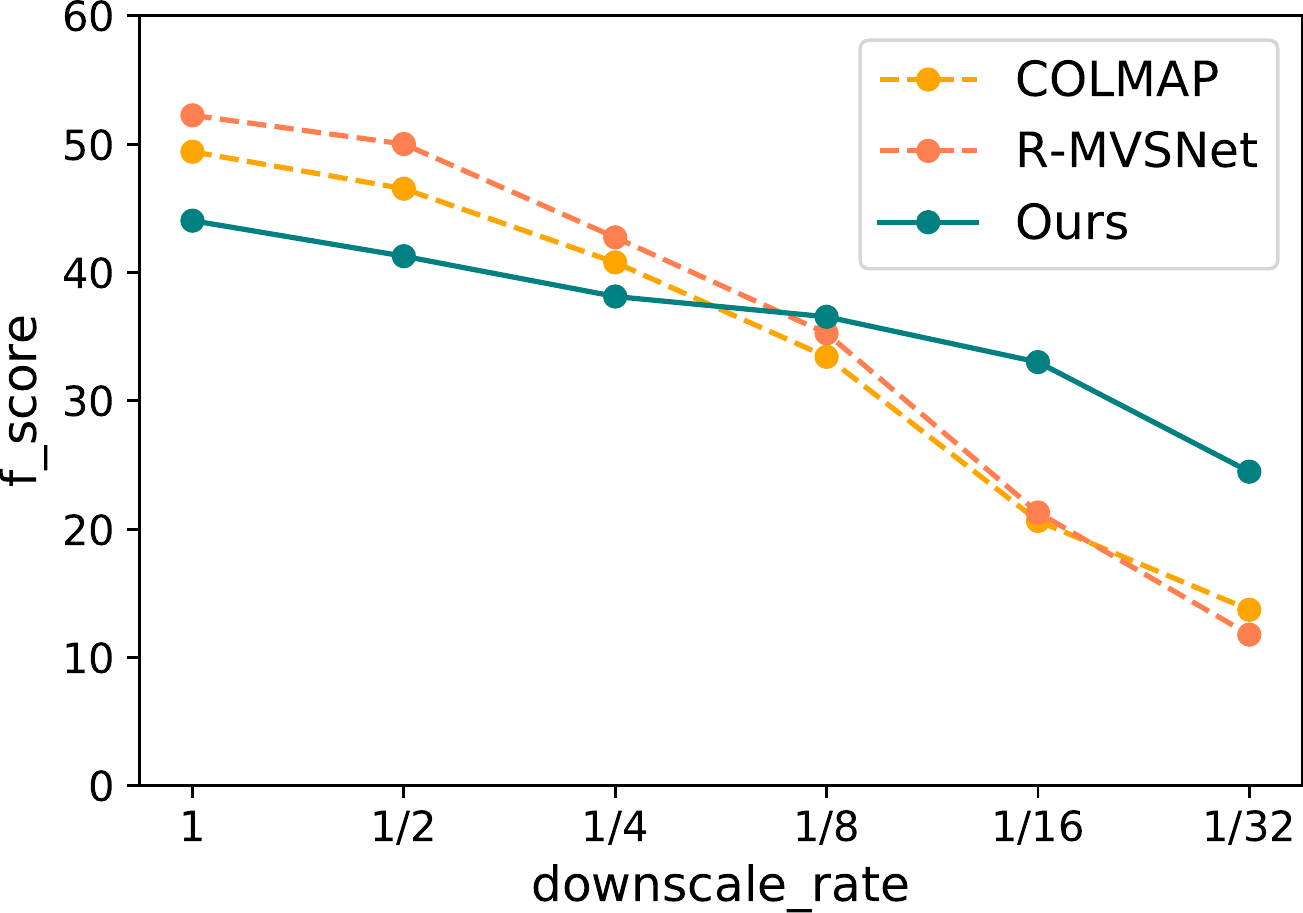}
}
\subfigure[temporary sub-sample]{                    
    \label{fig:tems}            
\includegraphics[width=54mm]{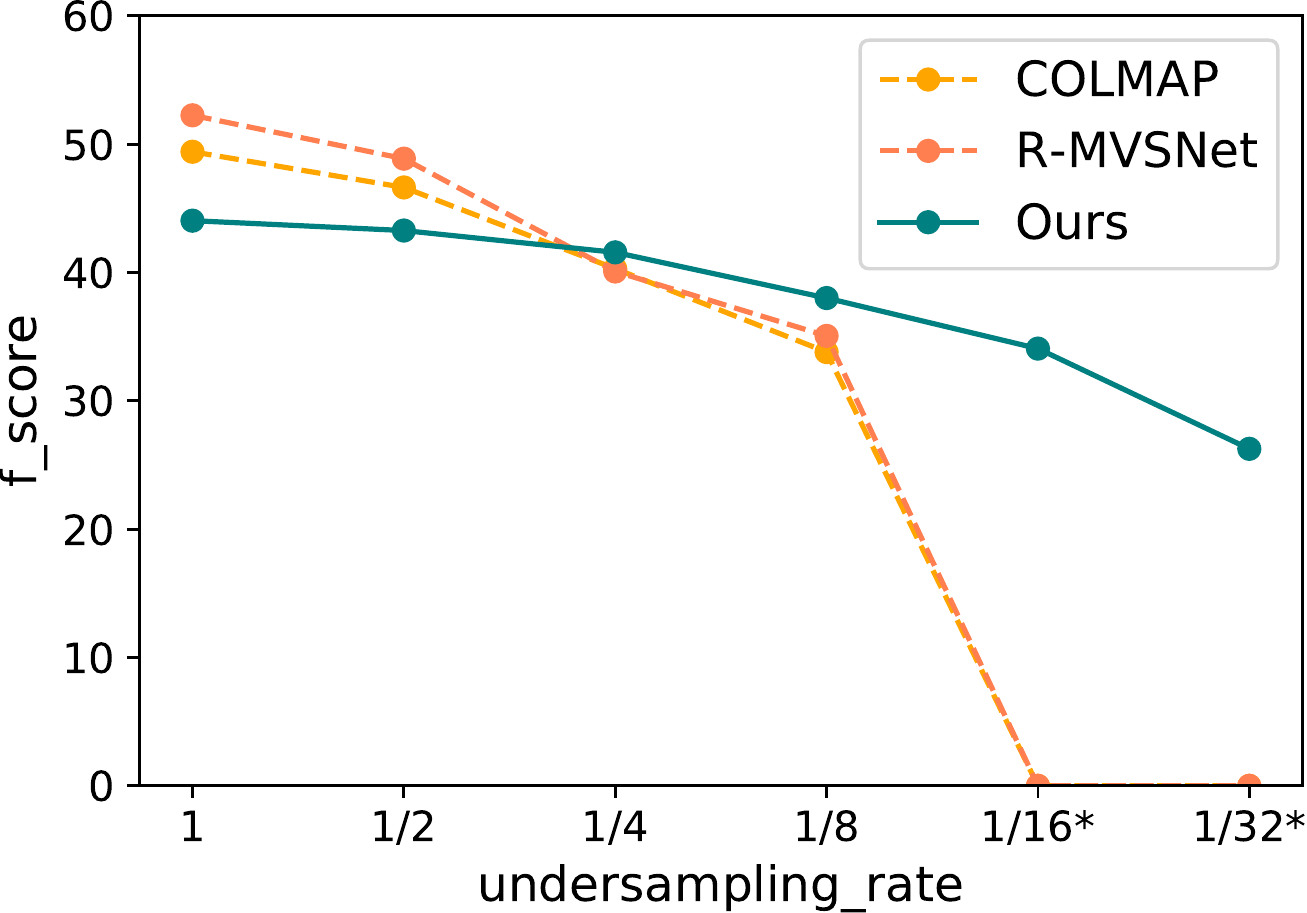}
}
\caption{Comparison with COLMAP\cite{schonberger2016structure} and R-MVSNet\cite{yao2019recurrent} with noisy input. Our work is less sensitive to initialization.}
\label{fig:noise} 
\end{figure}

\textbf{Tanks and Temples}
To evaluate the robustness to initialization quality, we compare DeepSFM with COLMAP and the SOTA -- R-MVSNet\cite{yao2019recurrent} on the Tanks and Temples\cite{Knapitsch2017} dataset as it contains densely captured high resolution images from which pose can be precisely estimated. 
To add noise on pose, we downscale the images and sub-sample temporal frames. For evaluation metrics, we adopt the F-score (higher is better) used in this dataset. The reconstruction qualities of Barn sequence are shown in fig.\ref{fig:noise}. 
It is observed that the performance of R-MVSNet and COLMAP drops significantly as the input quality becomes lower, while our method maintains the performance in a certain range. It is worth noting that COLMAP completely fails when the number of images are sub-sampled to 1/16.

\subsection{Model Analysis}
In this section, we analyze our model on several aspects to verify the optimality and show advantages over previous methods. {More ablation studies are provided in the supplementary material.}
\begin{figure}[tb]\centering      
\subfigure[depth metrics comparison]{                                   \label{fig:trans}           
\includegraphics[scale=0.34]{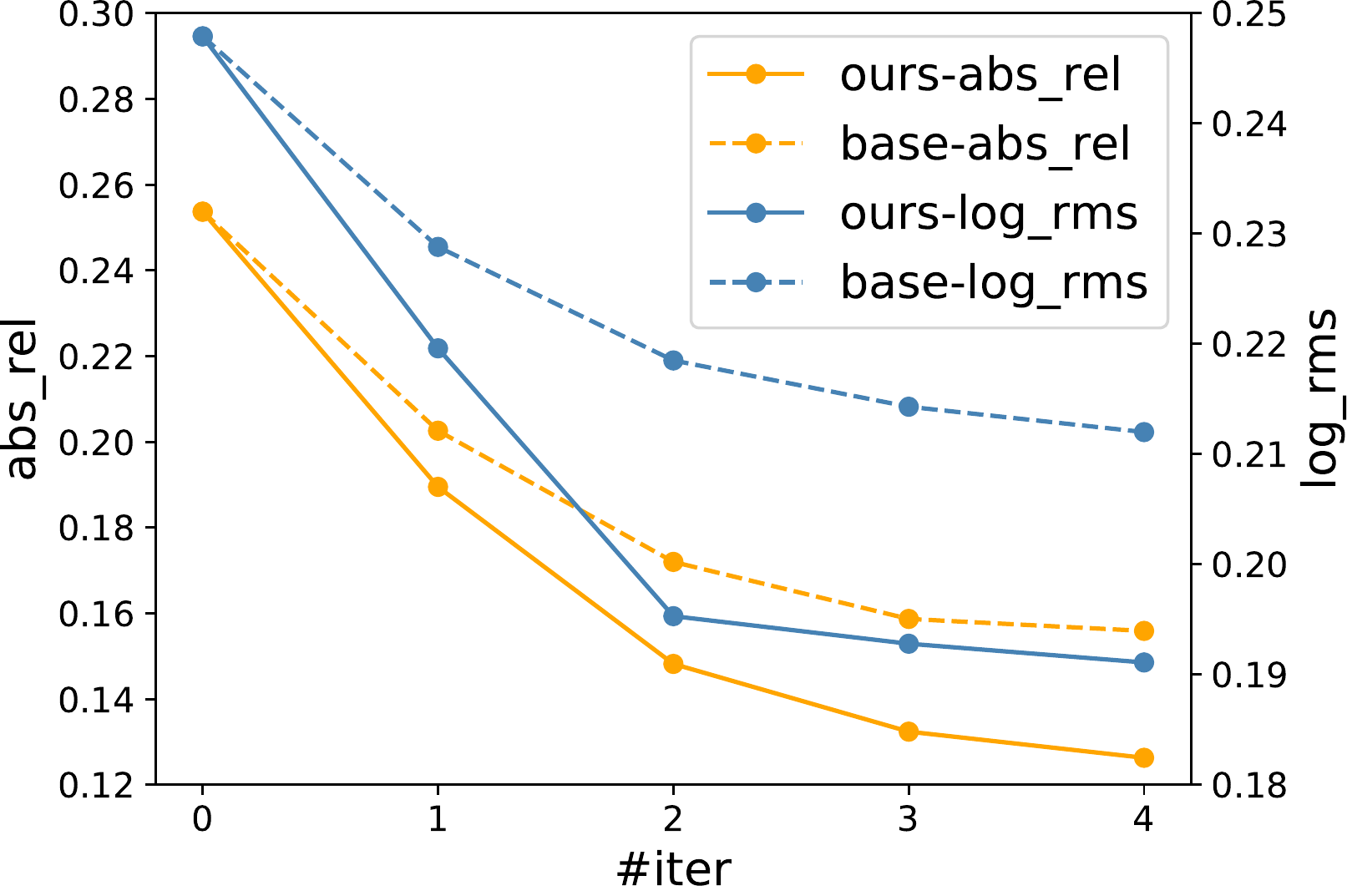}
}
\subfigure[camera pose metrics comparison]{                    
\label{fig:poses}            
\includegraphics[scale=0.34]{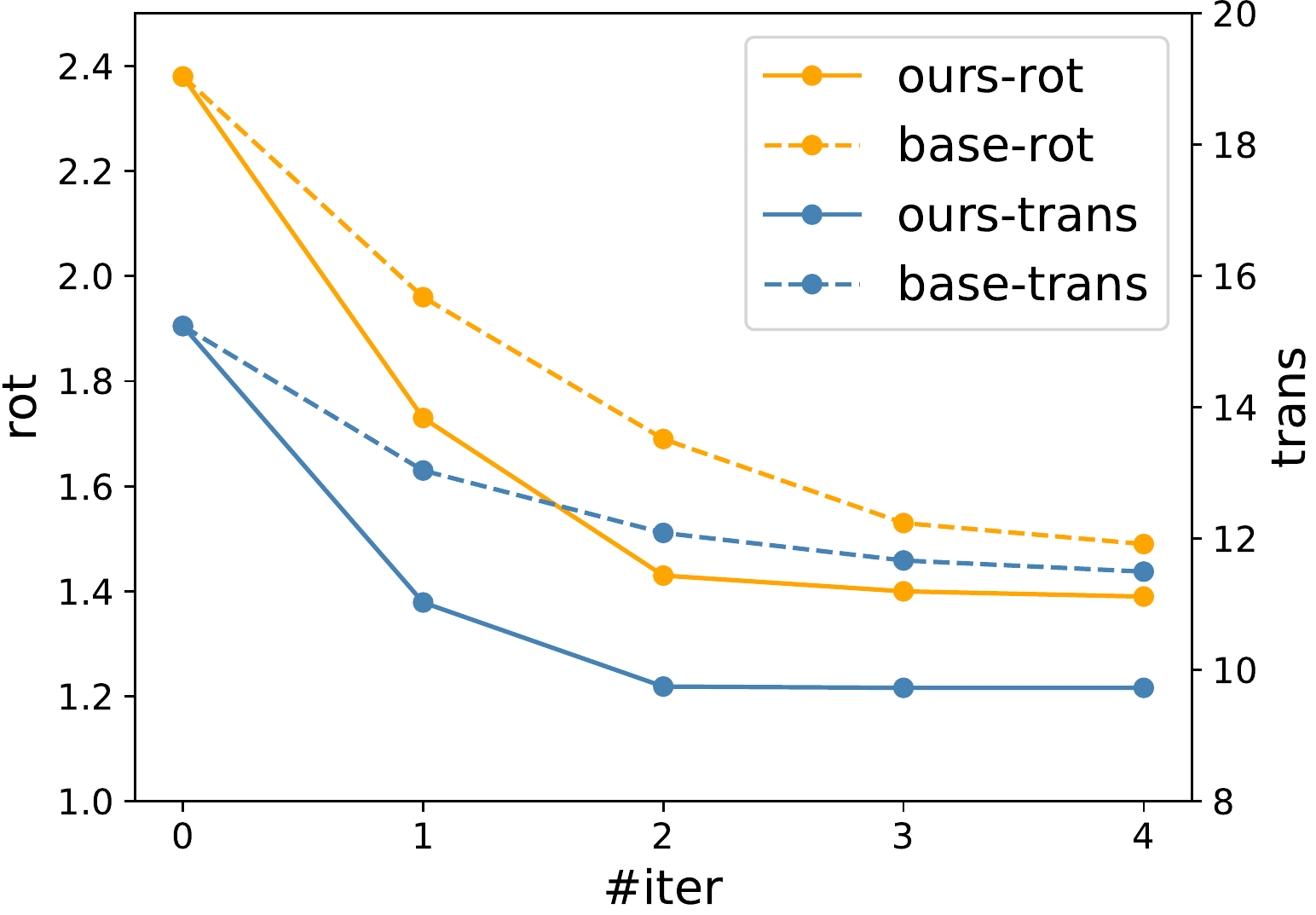}
}
\caption{Comparison with baseline during iterations. Our work converges at a better position. (a) abs relative error and log RMSE. (b) rotation and translation error.}
\label{fig:iteration} 
\end{figure}

\textbf{Iterative Improvement } Our model can run iteratively to reduce the prediction error. Fig.\ref{fig:iteration} (solid lines) shows our performance over iterations when initialized with the prediction from DeMoN.
As can be seen, our model effectively reduces both depth and pose errors upon the DeMoN output.
Throughout the iterations, better depth and pose benefit each other by building more accurate cost volume, and both are consistently improved.
The whole process is similar to coordinate descent algorithm, and finally converges at iteration 4.

\textbf{Effect of P-CV \label{sec:effect_pcv}} We compare DeepSFM to a baseline method for our P-CV. In this baseline, the depth prediction is the same as DeepSFM, but the pose prediction network is replaced by a direct visual odometry model \cite{steinbrucker2011real}, which updates camera parameters by minimizing pixel-wise photometric error between image features. 
Both methods are initialized with DeMoN results. As provided in Fig.\ref{fig:iteration}, DeepSFM consistently produces lower errors on both depth and pose over all the iterations. This shows that our P-CV predicts more accurate pose and performs more robust against noise depth at early stages. Fig. \ref{fig:eth3d_pose}(c) shows the visualized pose trajectories which are estimated by baseline(cyan) and our method(yellow) on ETH3D.

\begin{figure}[tb]\centering                                                          \subfigure[Abs relative error comparison]{                    \label{fig:multiviewcomp}             \includegraphics[scale=0.33]{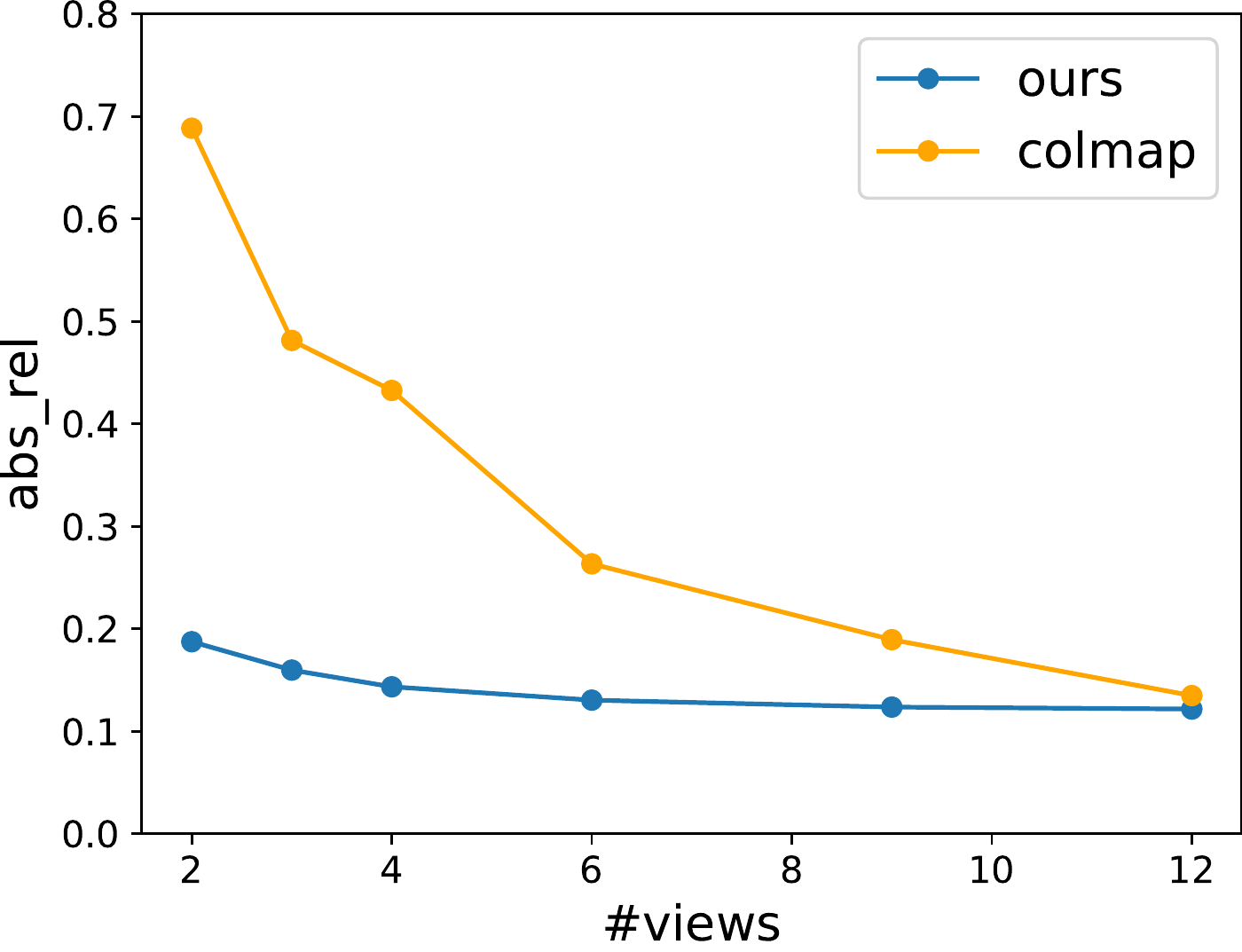}
}
\subfigure[Our results with different view numbers]{                                   \label{fig:multiviewvis}           \includegraphics[scale=0.32]{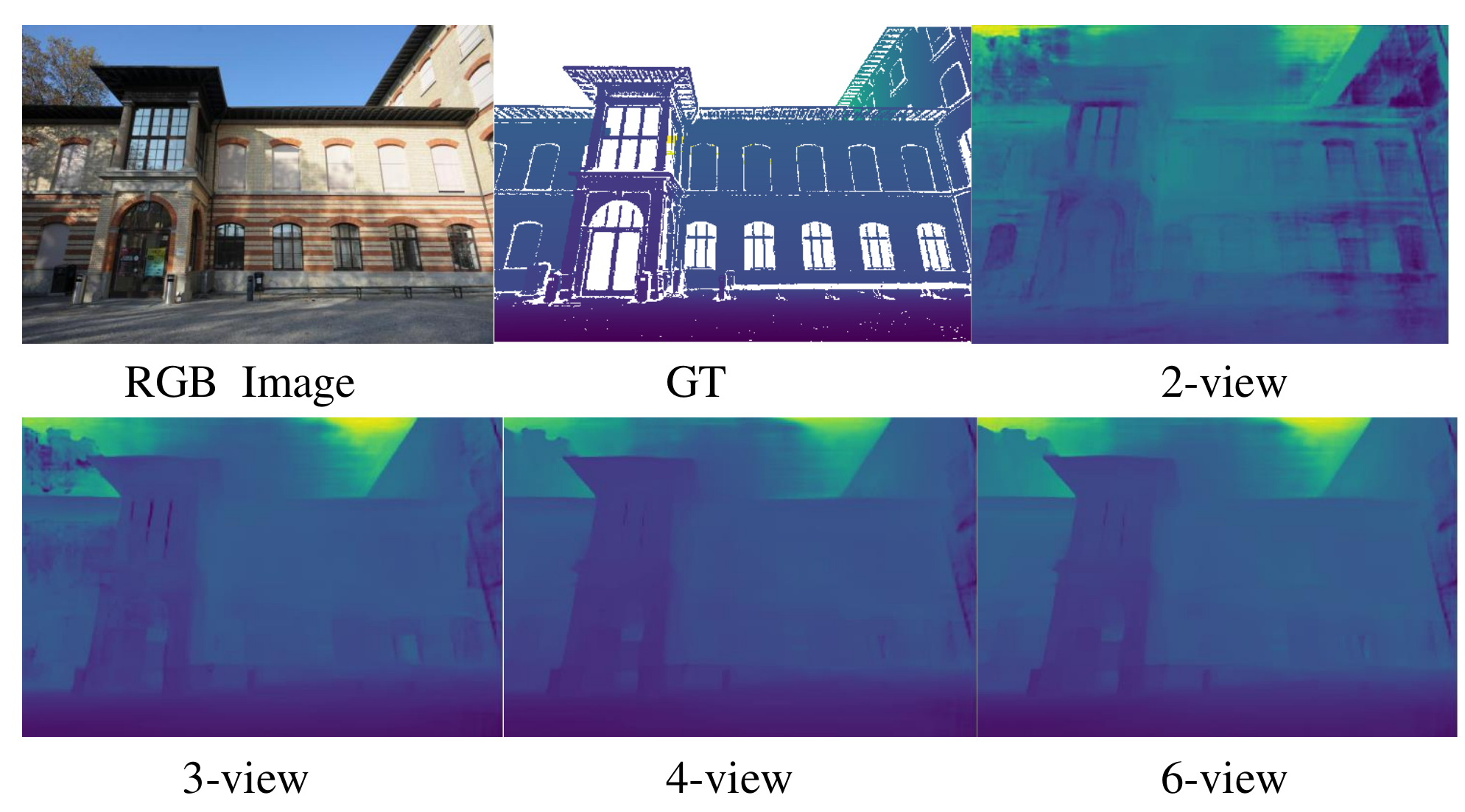}
}
\caption{Depth map results w.r.t. the number of images. Our performance does not change much with varying number of views. }
\label{fig:multiview} 
\end{figure}

\textbf{View Number} 
DeepSFM works still reasonably well with fewer views due to the free from optimization based components.
To show this, we compare to COLMAP with respect to the number of input views on ETH3D. As depicted in Fig.\ref{fig:multiview}, more images yield better results for both methods as expected.
However, our performance drops significantly slower than COLMAP with fewer number of inputs.
Numerically, DeepSFM cuts the depth error by half under the same number of views as COLMAP, or achieves similar error with half number of views required by COLMAP.
This clearly demonstrates that DeepSFM is more robust when fewer inputs are available.

\section{Conclusions}
\label{con}
We present a deep learning framework for Structure-from-Motion, which explicitly enforces photo-metric consistency, geometric consistency and camera motion constraints all in the deep network. 
This is achieved by two key components - namely D-CV and P-CV. Both cost volumes measure the photo-metric and geometric errors by hypothetically moving reconstructed scene points (structure) or camera (motion) respectively. Our deep network can be considered as an enhanced learning based BA algorithm, which takes the best benefits from both learnable priors and geometric rules. Consequently, our method outperforms conventional BA and state-of-the-art deep learning based methods for SfM.

\section*{Acknowledgements}
This project is partly supported by NSFC Projects (61702108), STCSM Projects (19511120700, and 19ZR1471800), SMSTM Project (2018SHZDZX01), SRIF Program (17DZ2260900), and ZJLab.

\bibliographystyle{splncs04}
\bibliography{0840}


\pagebreak

\begin{center}
\textbf{\Large Supplemental Materials}
\end{center}

\setcounter{section}{0}

\begin{figure*}[b]
\centering \includegraphics[width=122mm]{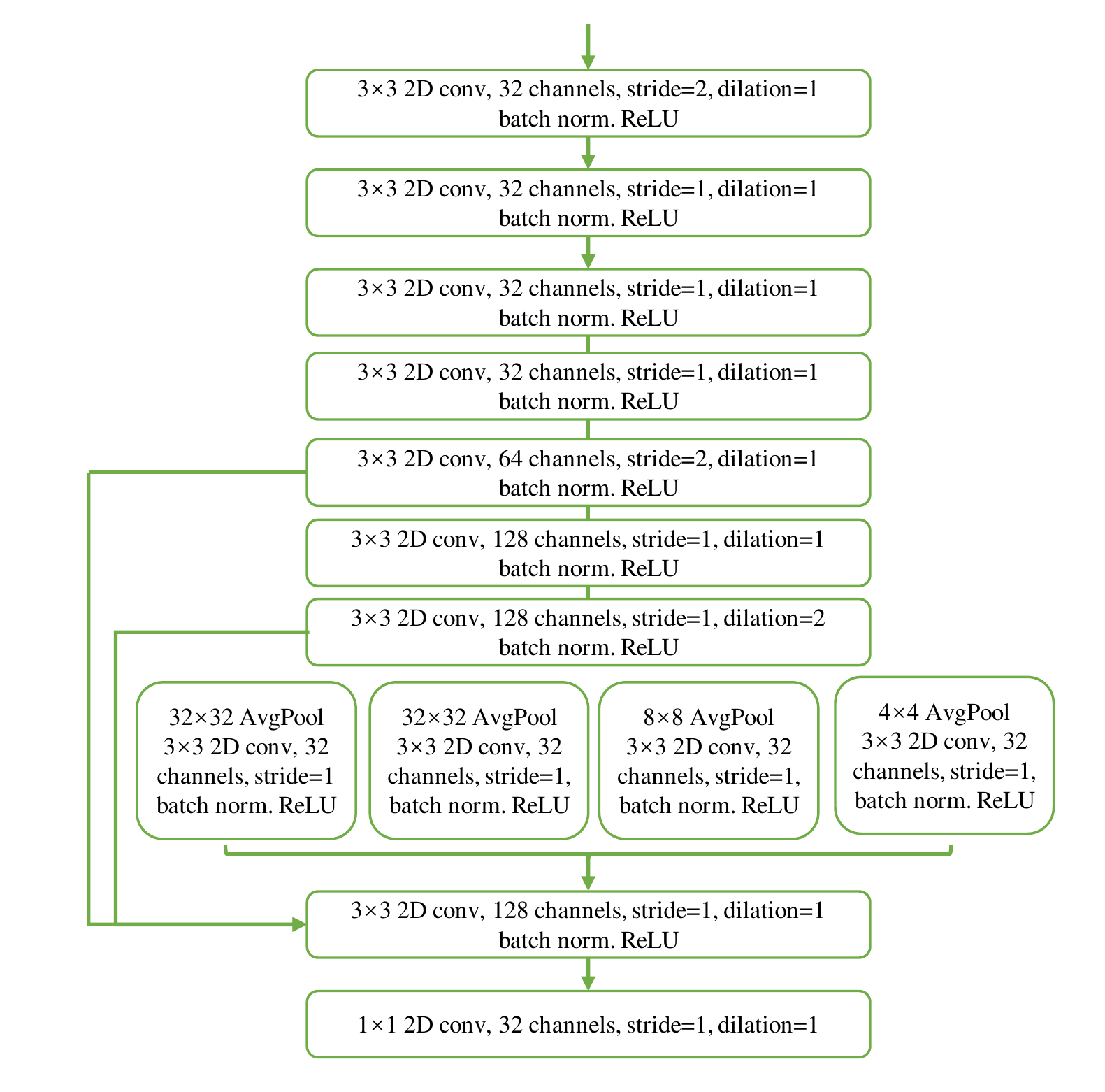}
\caption{Detail architecture of feature extractor.}
\label{fig:feature_extractor_detail} 
\end{figure*}

\begin{figure*}[tb]
\centering \includegraphics[width=122mm]{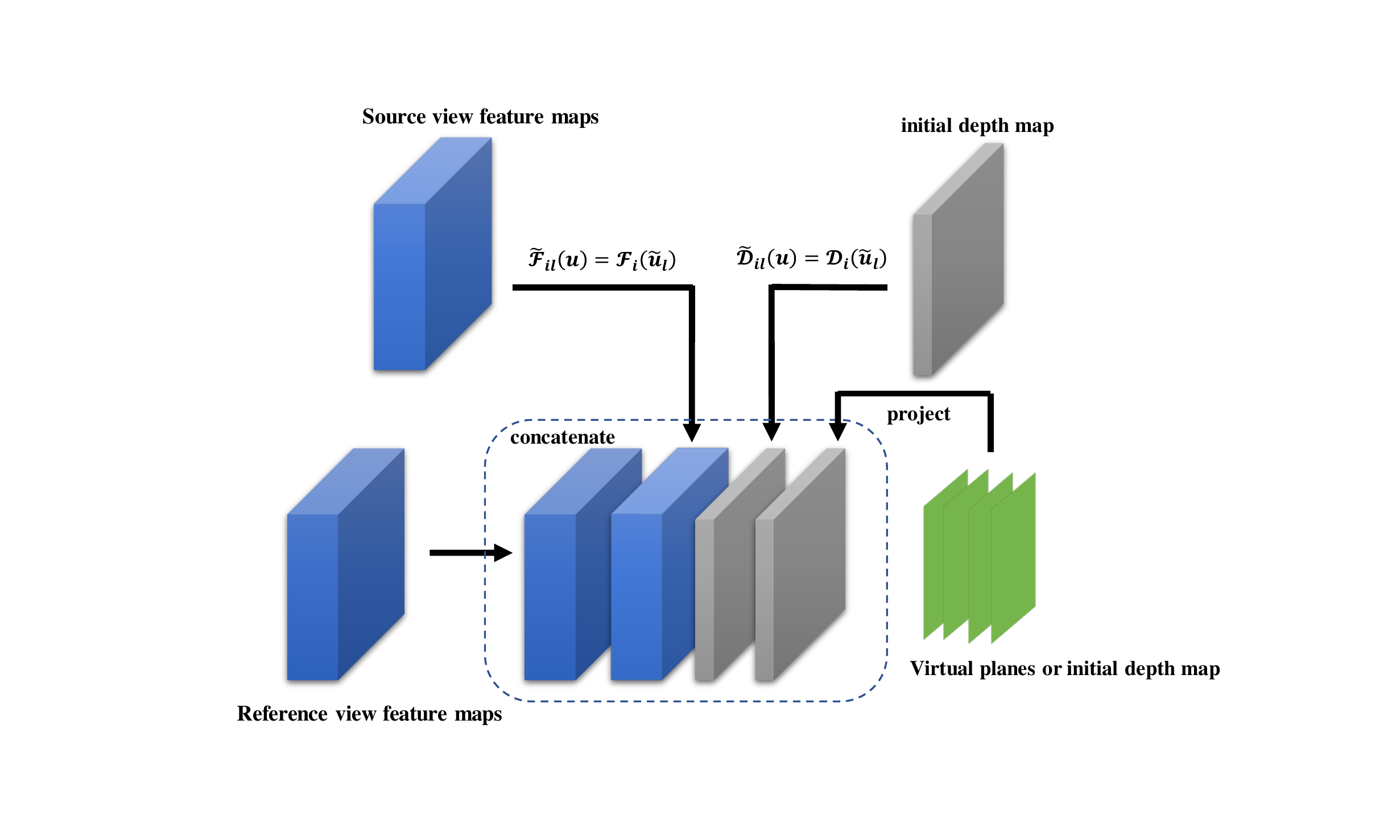}
\caption{Four components in D-CV or P-CV.}
\label{fig:cost_volume} 
\end{figure*}

\begin{figure*}[tb]
\centering \includegraphics[width=122mm]{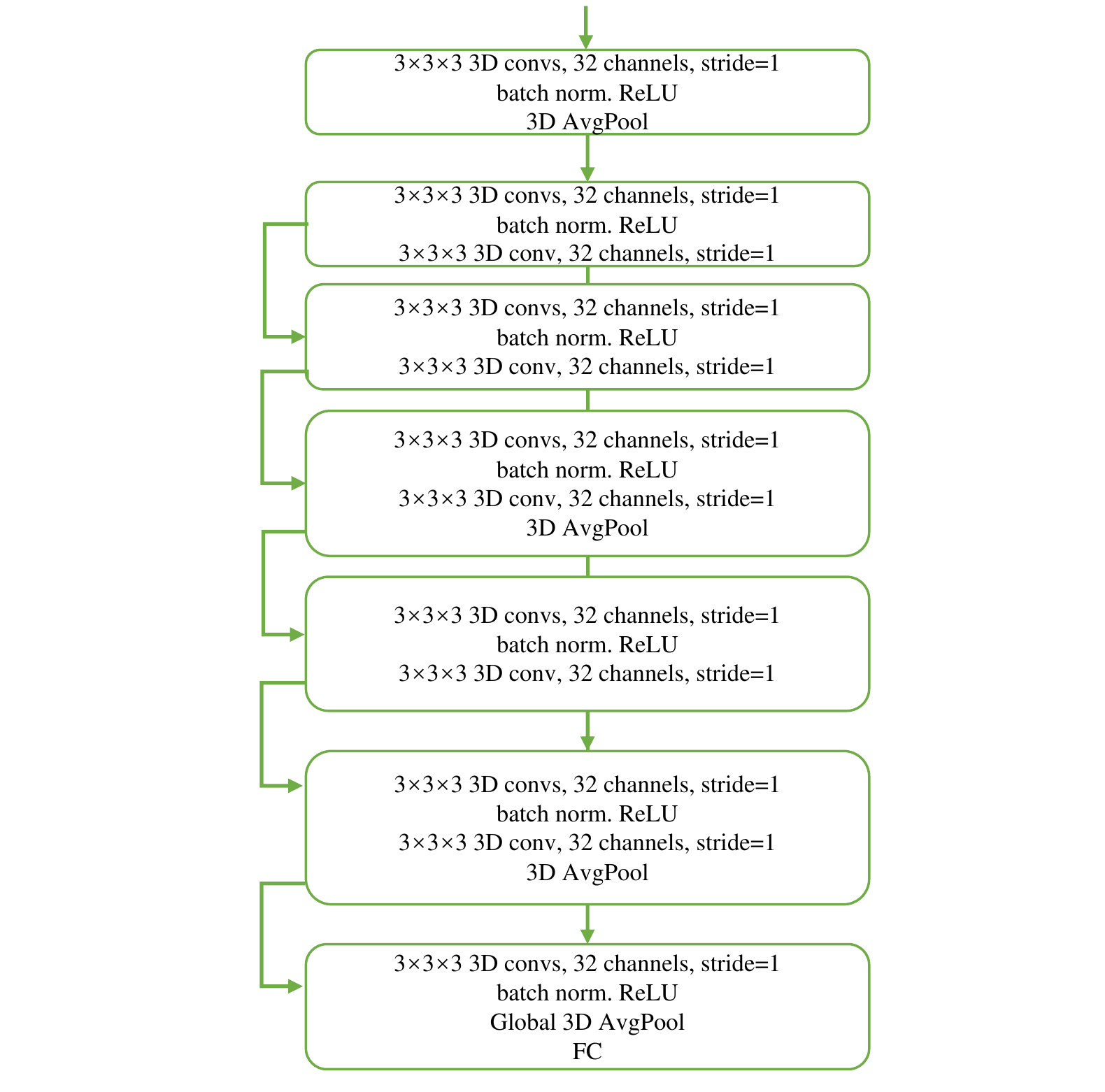}
\caption{3D convolutional layers After P-CV.}
\label{fig:3d_conv} 
\end{figure*}

\section{Implementation Details}
{We implement our system using PyTorch. The training procedure takes 6 days on 3 NVIDIA TITAN GPUs to converge on all 160k training sequences. The training batch size is set to 4, and the Adam optimizer ($\beta_1 = 0.9, \beta_2 = 0.999$) is used with learning rate $2\times10^{-4}$, which decreases to $4\times10^{-5}$ after 2 epochs. Within the first two epochs, the parameters in 2D CNN feature extraction module are initialized with pre-trained weights of \cite{im2018dpsnet} and frozen, and the ground truth depth maps for source images are used to construct D-CV and P-CV, which are replaced by predicted depth from network in latter epochs. During training, the length of input sequences is 2 (one target image and one source image). The sample number $L$ for D-CV is set to 64 and the sample number $P$ for P-CV is 1000. The range of both cost volumes is adapted during training and testing. For D-CV, its range is determined by the minimum depth values of the ground truth, which is the same as \cite{im2018dpsnet}. For P-CV, the bin size of rotation sampling is 0.035 and the bin size of translation sampling is $0.10 \times norm(t^*)$ for each initialization translation vector $t^*$.

\paragraph{Loss weights}
We follow two rules to set $\lambda_r$, $\lambda_t$ and $\lambda_d$ for $\mathcal{L}_{final}$: 1) the loss term provides gradient on the same order of numerical magnitude, such that no single loss term dominates the training process. This is because accuracy in depth and camera pose are both important to reach a good consensus. 2) we found in practice the camera rotation has higher impact on the accuracy of the depth but not the opposite. To encourage better performance of pose, we set a relatively large $\lambda_r.$ In practice, the weight parameter $\lambda$ for $\mathcal{L}_{depth}$ to balance loss objective is set to 0.7, while $\lambda_r=0.8$, $\lambda_t=0.1$ and $\lambda_d=0.1$. }

\paragraph{Feature extraction module}
As shown in Fig.\ref{fig:feature_extractor_detail}, we build our feature extraction module referring to DPSNet \cite{im2018dpsnet}. The module takes $4W \times 4H \times 3$ images as input and output feature maps of size $W \times H \times 32$, which are used to build D-CV and P-CV.

\paragraph{Cost volumes}
Figure \ref{fig:cost_volume} shows the detailed components for the P-CV and D-CV. Each channel of cost volume is composed of four components: reference view feature maps, warped source view feature maps, the warped source view initial depth map and the projected reference view depth plane or initial depth map. For P-CV construction, we take each sampled hypothetical camera pose, and carry out the warping process on source view depth maps and initial depth map based on the camera pose. And the initial reference view depth map is projected to align numeric values with the warped source view depth map. Finally those four components are concatenated as one channel of 4D P-CV. We do this on all P sampled camera poses, and get the P channel P-CV. The building approach for D-CV is similar, we take each sampled hypothetical depth plane, and carry out warping process on source view feature maps and the initial depth map. And the depth plane is projected to align with the source view depth map. After concatenation, one channel in D-CV is got. Same computation is done based on all L virtual depth planes, and the L channel D-CV is built up.

\paragraph{3D convolutional layers}
The detail architecture of 3D convolutional layers after D-CV is almost the same as DPSNet \cite{im2018dpsnet}, except for the fist convolution layer. In order to compatible with the newly introduced depth consistent components in D-CV, We adjust the input channel number to 66 instead of 64. As shown in Fig.\ref{fig:3d_conv}, for 3D convolutional layers after P-CV, the architecture is similar to D-CV 3D convolution layers with three extra 3D average pooling layers and finally there is one global average pooling in the dimensions of image width and height, after which we get a $P \times 1 \times 1$ tensor.

\section{Evaluation on ScanNet}

ScanNet\cite{dai2017scannet} provides a large set of indoor sequences with camera poses and depth maps captured from a commodity RGBD sensor. Following BA-Net\cite{tang2018ba}, we leverage this dataset to evaluate the generalization capability by training models on DeMoN and testing here. The testing set is the same as BA-Net, which takes 2000 pairs filtered from 100 sequences.

We evaluate the generalization capability of DeepSFM on ScanNet. Table \ref{tab:scannet} shows the quantitative evaluation results for models trained on DeMoN. The results of BA-Net, DeMoN\cite{ummenhofer2017demon}, LSD-SLAM\cite{engel2014lsd} and Geometric BA\cite{nister2004efficient} are obtained from \cite{tang2018ba}. 
As can be seen, our method significantly outperforms all previous work, which indicates that our model generalizes well to general indoor environments.
\begin{table*}
\begin{center}
\caption{\label{tab:scannet}Results on ScanNet. (sc\_inv: scale invariant log rms; \textbf{Bold}: best.)}
\begin{tabular}{cccccccc}
\toprule[2px] 
{Method} & \multicolumn{5}{c}{{Depth}} & \multicolumn{2}{c}{{Motion}}\tabularnewline
\cmidrule{2-8} \cmidrule{3-8} \cmidrule{4-8} \cmidrule{5-8} \cmidrule{6-8} \cmidrule{7-8} \cmidrule{8-8} 
 & {abs\_rel} & {sq\_rel} & {rms} & {log\_rms} & {sc\_inv} & {Rot} & {Trans}\tabularnewline
\midrule[2px]  
{Ours} & \textbf{0.227} & \textbf{0.170} & \textbf{0.479} & \textbf{0.271} & \textbf{0.268} & {1.588} & \textbf{30.613}\tabularnewline
{BA-Net} & {0.238} & {0.176} & 0.488 & {0.279} & {0.276} & \textbf{1.587} & {31.005}\tabularnewline
{DeMoN} & {0.231} & {0.520} & {0.761} & {0.289} & {0.284} & {3.791} & {31.626}\tabularnewline
{LSD-SLAM} & {0.268} & {0.427} & {0.788} & {0.330} & {0.323} & {4.409} & {34.360}\tabularnewline
{Geometric BA} & {0.382} & {1.163} & {0.876} & {0.366} & {0.357} & {8.560} & {39.392}\tabularnewline
\bottomrule[2px] 
\end{tabular}

\end{center}
\end{table*}
\begin{figure}[tb]\centering 

\subfigure[resolution downscale]{       \centering    
\label{fig:resd1}           
\includegraphics[width=36mm]{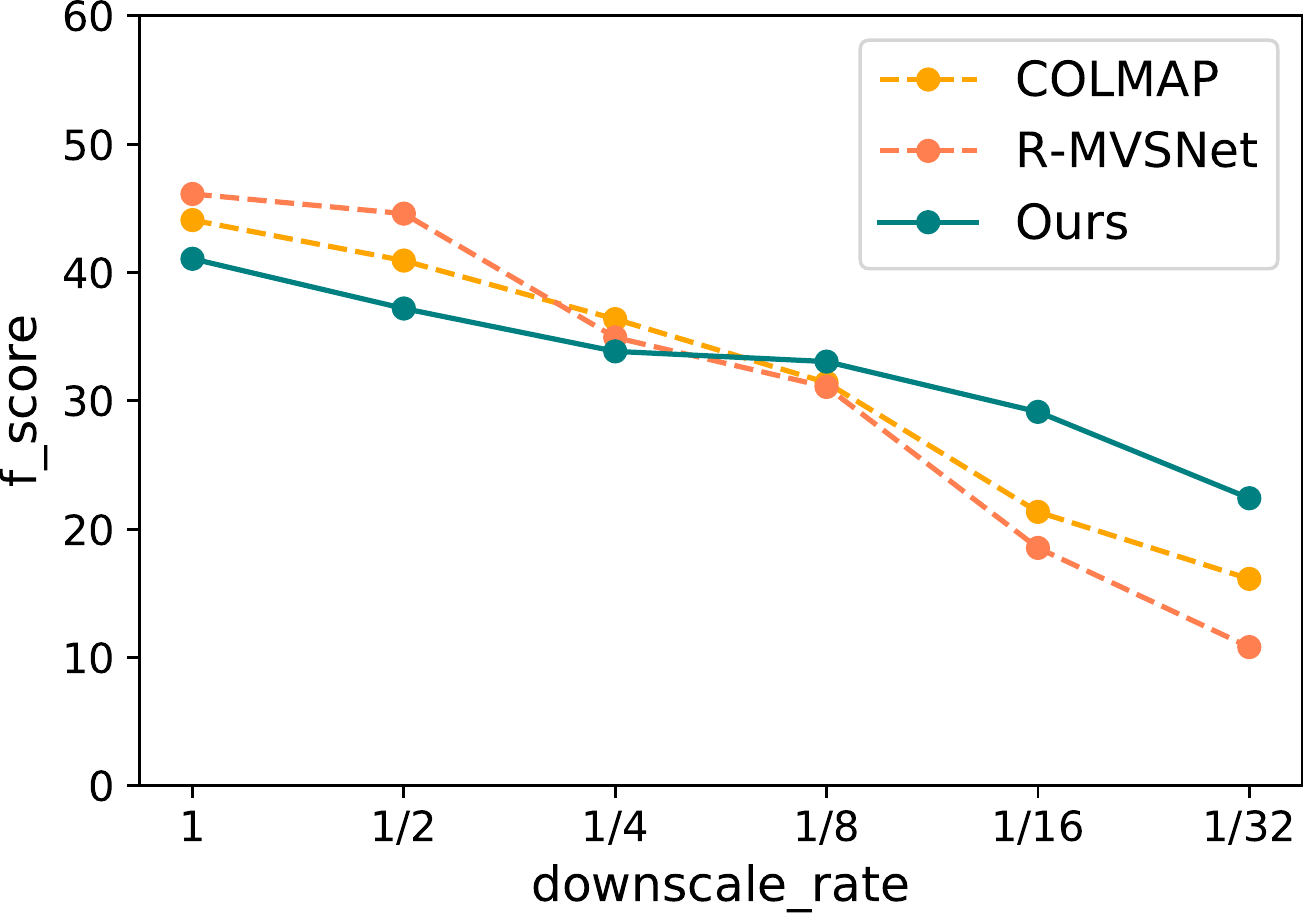}
}
\subfigure[temporary sub-sample]{                    
    \label{fig:tems1}            
\includegraphics[width=36mm]{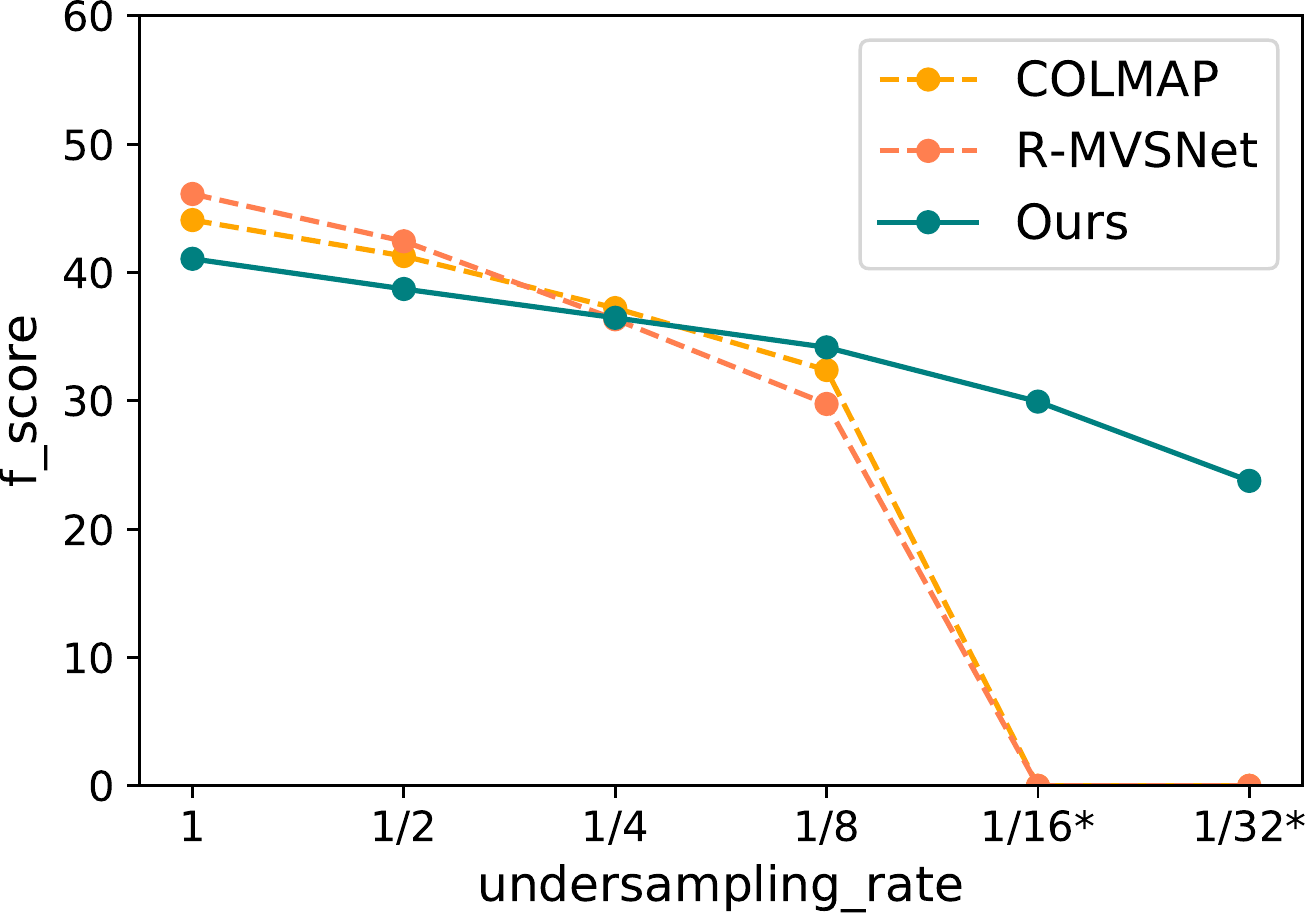}
}
\subfigure[Gaussian noise]{                    
    \label{fig:guass_n}            
\includegraphics[width=36mm]{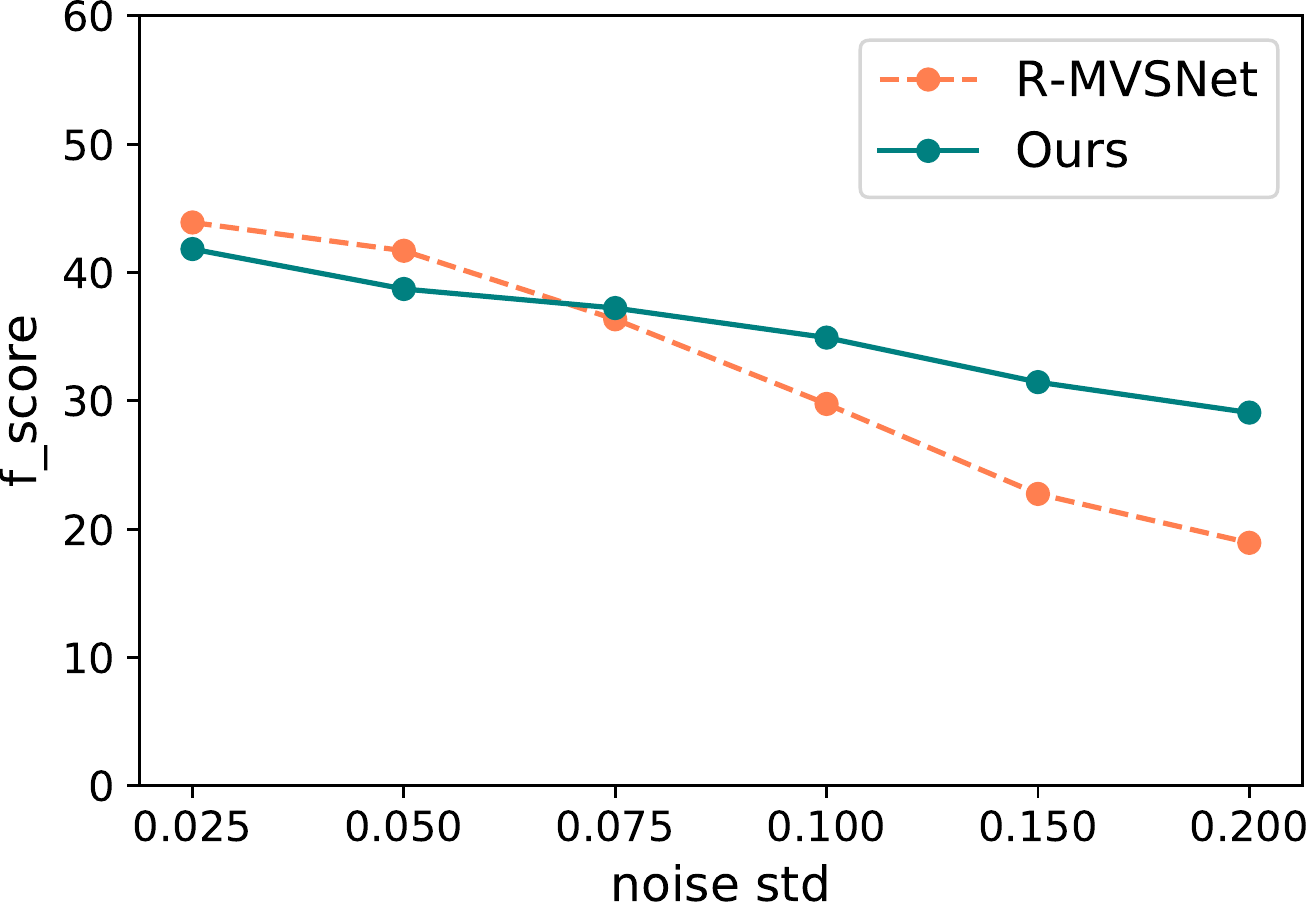}
}
\caption{Comparison with COLMAP\cite{schonberger2016structure} and R-MVSNet\cite{yao2019recurrent} with noisy input. Our work is less sensitive to initialization.}
\label{fig:noise_t} 
\end{figure}
\section{Evaluation on Tanks and Temples}
As illustrated in Section4.2, We compare DeepSFM with COLMAP and R-MVSNet\cite{yao2019recurrent} on the Tanks and Temples\cite{Knapitsch2017} dataset. Figure \ref{fig:noise_t} are more experimental results on Tanks and Temples dataset. All 7 training sequences provided by the dataset are used for the evaluation and the F-score are calculated as average. We add noise to COLMAP poses by down-scaling the images, sub-sampling temporal frames or directly add random Gaussian noise. Compared with COLMAP and R-MVSNet, our method is robuster to initialization quality.
\section{Computational costs}
{
The computational costs on DeMoN dataset are shown in Table. \ref{tab:comp_cost}. The memory cost of DeMoN and ours is the peak memory usage during testing on a TiTAN X GPU. 
}

\begin{table*}
\begin{center}
\caption{\label{tab:comp_cost}The computational costs on DeMoN dataset.}
\begin{tabular}{cccc}
\toprule[2px]  
Network&Ours&BANet&DeMoN\tabularnewline
\midrule[2px]  
Memory/image&1.17G&2.30G&0.60G\tabularnewline
Runtime/image&410ms&95ms&110ms\tabularnewline
Resolution&640*480&320*240&256*192\tabularnewline
\bottomrule[2px]
\end{tabular}

\end{center}
\end{table*}

\begin{table*}
\begin{center}
\caption{\label{tab:more_iter}The performance of the optimization iterations for testing.}
\begin{tabular}{ccccccc}
\toprule[2px]  
 & Initialization & Iteration 2 & Iteration 4 & Iteration 6 & Iteration 10 & Iteration 20\tabularnewline
\midrule[2px]  

abs relative & 0.254 & 0.153 & 0.126 & 0.121 & 0.120 & 0.120\tabularnewline

log rms & 0.248 & 0.195 & 0.191 & 0.190 & 0.190 & 0.191\tabularnewline

translation & 15.20 & 9.75 & 9.73 & 9.73 & 9.73 & 9.73\tabularnewline

rotation & 2.38 & 1.43 & 1.40 & 1.39 & 1.39 & 1.39\tabularnewline
\bottomrule[2px]
\end{tabular}

\end{center}
\end{table*}
\begin{table*}
\begin{center}
\caption{\label{tab:warping_func}The performance with different warping methods.}
\begin{tabular}{cccccc}
\toprule[2px]  
MVS Dataset & L1-inv & sc-inv & L1-rel & Rot & Trans\tabularnewline
\midrule[2px]  

Billinear interpolation & 0.023 & 0.134 & 0.079 & 2.867 & 9.910\tabularnewline

Nearest neighbor & 0.021 & 0.129 & 0.076 & 2.824 & 9.881\tabularnewline
\bottomrule[2px] 
\end{tabular}

\end{center}
\end{table*}
\section{More Ablation Study}
\subsection{More Iterations for Testing} 
We take up to four iterations when we train DeepSFM. During inference, the predicted depth maps and camera poses of previous iteration are taken as initialization of next iteration. To show how DeepSFM performs with more iterations than it is trained with, we show results in Table \ref{tab:more_iter}. We tested with up to 20 iterations, and it converges at the 6-th iteration.

\subsection{Bilinear Interpolation vs Nearest Neighbor Sampling} 
For the construction of D-CV and P-CV, depth maps are warped via the nearest neighbor sampling instead of bilinear interpolation. Due to the discontinuity of the depth values in depth maps, the bilinear interpolation may bring some side effects. It may do damage to the geometry consistency and smooth the depth boundaries. As a comparison, we replace the nearest neighbor sampling with the bilinear interpolation. As shown in Table \ref{tab:warping_func}, the performance of our model gains a slight drop with the bilinear interpolation, which indicates that the nearest neighbor sampling method is indeed more geometrically meaningful for depth. In contrast, the differentiable bilinear interpolation is required for the warping of image features, whose gradients are back propagated to feature extractor layers. Further exploration will be an interesting future work.

{ \subsection{Geometric consistency}
We include both the image features and the initial depth values into the cost volumes to enforce photo-consistency and geometric consistency. To validate the geometric consistency, we conduct an ablation study on MVS dataset and show the depth accuracy w/ and w/o geometric consistency with same GT poses in Table \ref{tab:geo_con}. Meanwhile, as shown in Fig. \ref{fig:vis_e1}, the geometric consistency is especially helpful for regions with weak photometric consistency, e.g. textureless, specular reflection.

\begin{table*}
\begin{center}
\caption{\label{tab:geo_con}Results on MVS with and w/o geometric consistency($\alpha=1.25$). The metrics are the same as those on Eth3D datasets. }
\begin{tabular}{ccccccccc}
\toprule[2px] 
Method &  abs\_rel  &  abs\_diff  &  sq\_rel  &  rms  &  log\_rms  &  $\delta<\alpha$  &   $\delta<\alpha^2$  &  $ \delta<\alpha^3 $  \tabularnewline
\midrule[2px]
w/  & 0.0698 & 0.1629 & 0.0523 & 0.3620 & 0.1392 & 90.25 & 96.06 & 98.18\tabularnewline
w/o & 0.0813 & 0.2006 & 0.0971 & 0.4419 &	0.1595 & 88.53 & 94.54 & 97.35\tabularnewline
\bottomrule[2px] 
\end{tabular}
\end{center}
\end{table*}
}
{\subsection{Initialization data augmentation} 
Adding random noises to the initialization is a commonly used way to increase the robustness of the pipeline. As a comparison, we initialize our pipeline by adding small Gaussian random noises to DeMoN pose and depth map results and then fine-tune the network on the training set of DeMoN datasets. After the fine-tuning, we test our method on MVS dataset on which the network is not trained on, and the performance of our network decreases slightly after the data augmentation. This demonstrates that adding small Gaussian random noises dose not increase the generalization ability of our method on unseen data, since it's easy for the network to over fit the noise distribution. 
} 

\begin{table*}
\begin{center}
\caption{\label{tab:smooth}Results on MVS with and w/o scale invariant gradient loss($\alpha=1.25$). The metrics are the same as those on Eth3D datasets. }
\begin{tabular}{ccccccccc}
\toprule[2px] 
Method &  abs\_rel  &  abs\_diff  &  sq\_rel  &  rms  &  log\_rms  &  $\delta<\alpha$  &   $\delta<\alpha^2$  &  $ \delta<\alpha^3 $  \tabularnewline
\midrule[2px]
w/  & 0.0712 & 0.1630 & 0.0531 & 0.3637 & 0.1379 & 90.25 & 96.02 & 98.24\tabularnewline
w/o & 0.0698 & 0.1629 & 0.0523 & 0.3620 & 0.1392 & 90.25 & 96.06 & 98.18\tabularnewline
\bottomrule[2px] 
\end{tabular}
\end{center}
\end{table*}
\begin{figure*}[tb]
\centering \includegraphics[width=122mm]{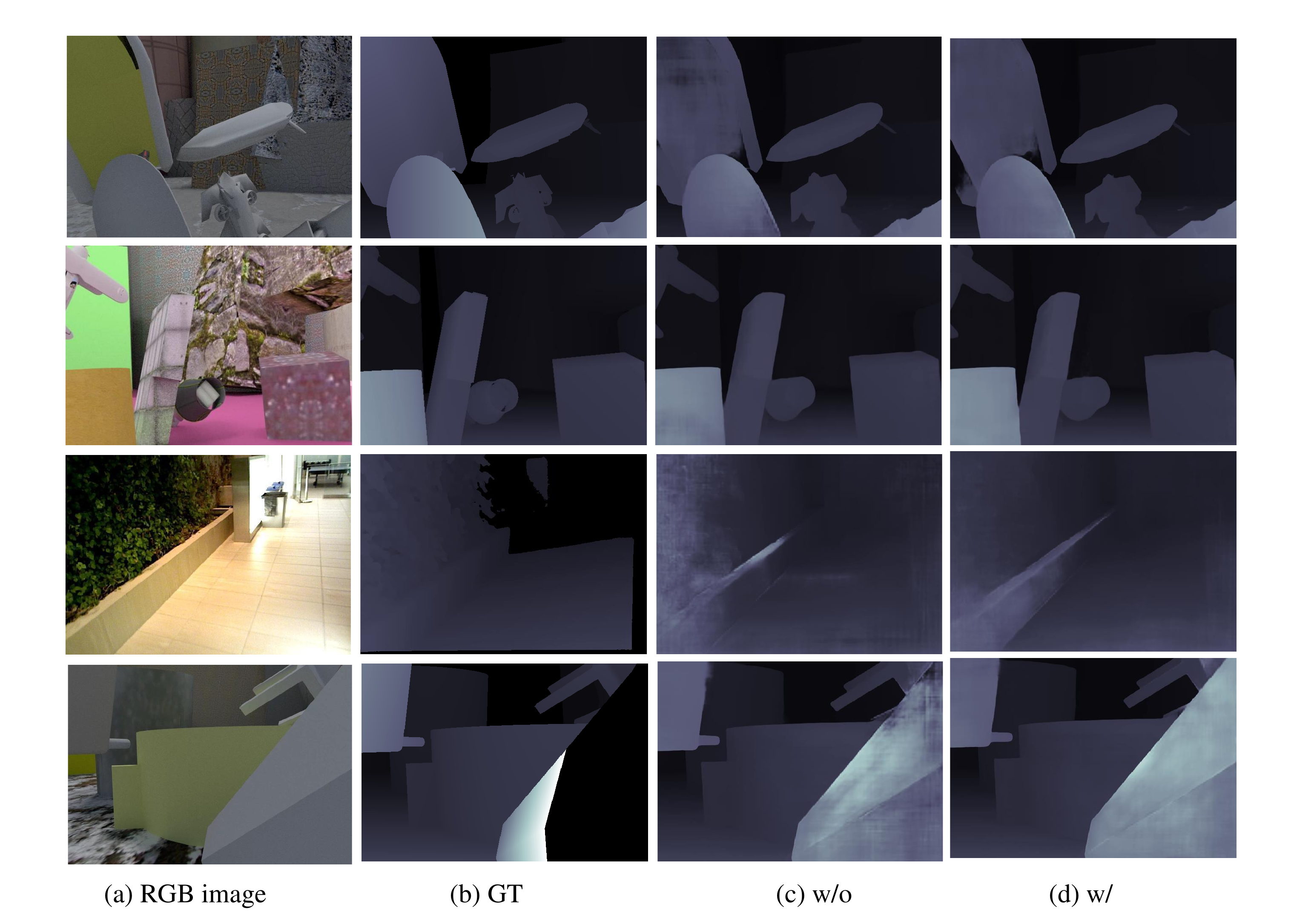}
\caption{ Qualitative comparisons with and w/o scale invariant gradient loss.}
\label{fig:smoo} \end{figure*}
{ \subsection{Smoothness on depth map} 
When compared with DeMoN\cite{ummenhofer2017demon}, the output depth maps of ours are sometimes less spatially smooth. Besides L1 loss on the inverse depth map values, DeMoN applied scale invariant gradient loss for depth supervision, which enhances the smoothness of estimated depth maps. To address the smoothness issue, we add scale invariant gradient loss and set its weight as 1.5 times of L1 loss follow DeMoN to retrain the network. As shown in Table \ref{tab:smooth}, no significant improvement is observed on depth evaluation metrics. 
Nevertheless, there are qualitative improvements of depth map in some samples as shown in Fig. \ref{fig:smoo}.
}
\begin{table*}
\begin{center}
\caption{\label{tab:sample}Results on MVS with different bin size (first column) of rotation/translation. The metrics are the same as those on DeMoN datasets. }
\begin{tabular}{ccc|ccc}
\toprule[2px] 
Rotation(radian) & Rot error& Trans error & Translation($\times$norm) & Rot error & Trans error\tabularnewline
\midrule[2px]
0.07 & 3.024  & 9.974 & 0.20& 3.252 & 10.117\tabularnewline

0.05 & 2.916 & 9.890 & 0.15 & 3.080 & 9.758\tabularnewline

0.03 & 2.825 & 9.836 & 0.10 & 2.825 & 9.836\tabularnewline

0.02 & 2.893 & 9.941 & 0.05 & 3.308 & 11.013\tabularnewline
\bottomrule[2px] 
\end{tabular}
\end{center}
\end{table*}
{\subsection{Pose sampling} 
 As described in section 3.3 of the paper, We use the same strategy for pose space sampling on different datasets. To show the generalization and the robustness of our method with different pose sampling strategies, we show the performance of our method with different bin size of rotation/translation without retraining in Table \ref{tab:sample}. Our model is a fully physical-driven
architecture and shows well generalization ability.
}

{\section{Discussion with DeepV2D}
DeepV2D\cite{teed2018deepv2d} is a concurrent learning-based SfM method, which has shown excellent performance across a variety of datasets and tasks. We couldn't make a fair comparison with DeepV2D due to different settings of two methods. Here is a brief discussion and comparison. DeepV2D composes geometrical algorithms into the differentiable motion module and the depth module, and updates depth and camera motion alternatively. The depth module of DeepV2D builds a cost volume which is similar to our work except for the geometric consistency introduced by our method. The motion module of DeepV2D minimizes the photometric re-projection error between image features of each pair via Gauss-Newton iterations, while our method learns correspondence of photometric and geometric features between each pair by P-CV and 3D conv. 

}

\section{Visualization}

\begin{figure*}[tb]
\centering \includegraphics[width=122mm]{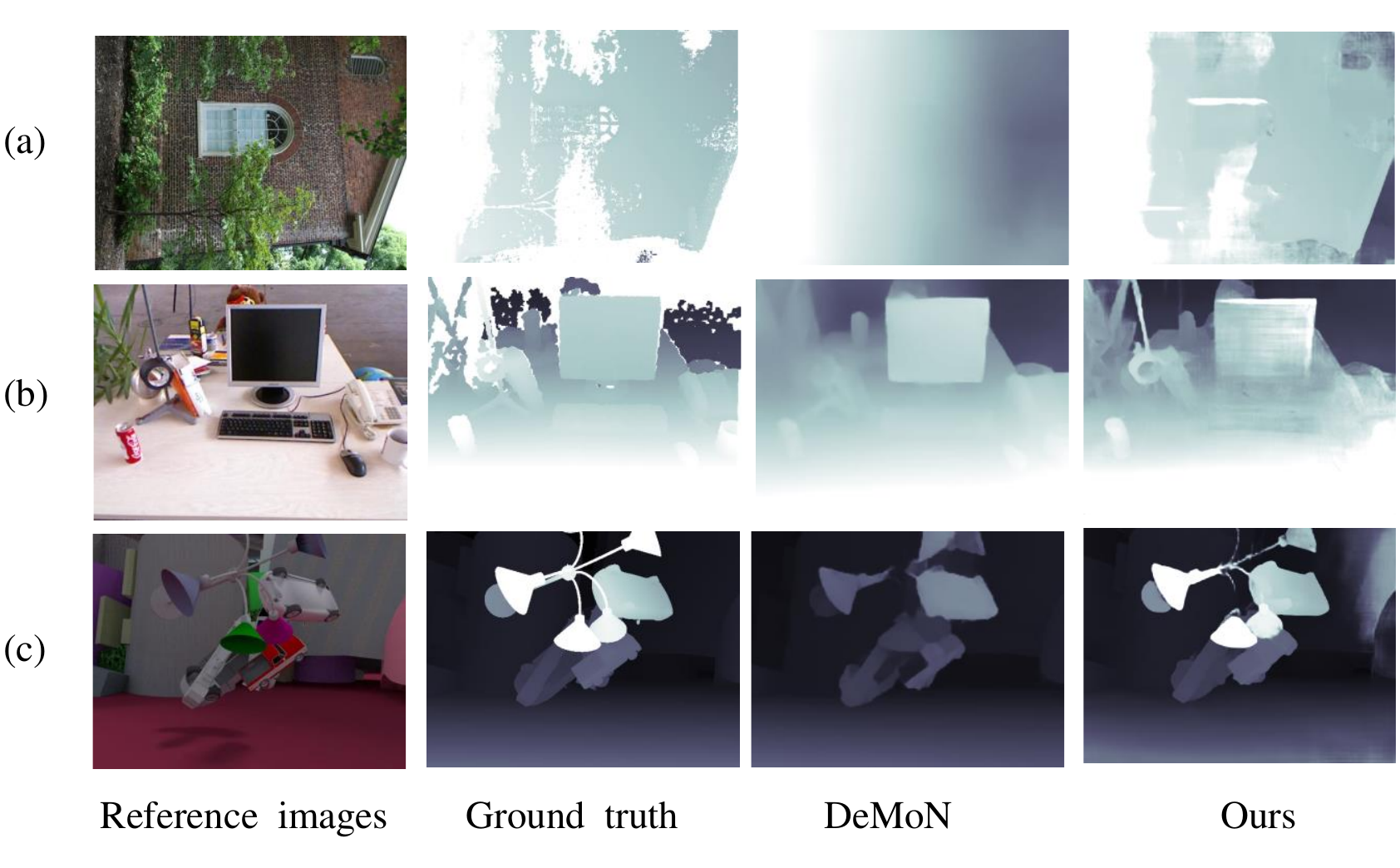}
\caption{More Qualitative comparisons with DeMoN \cite{ummenhofer2017demon} on DeMoN datasets.}
\label{fig:vis_d} 
\end{figure*}

\begin{figure*}[tb]
\centering \includegraphics[width=122mm]{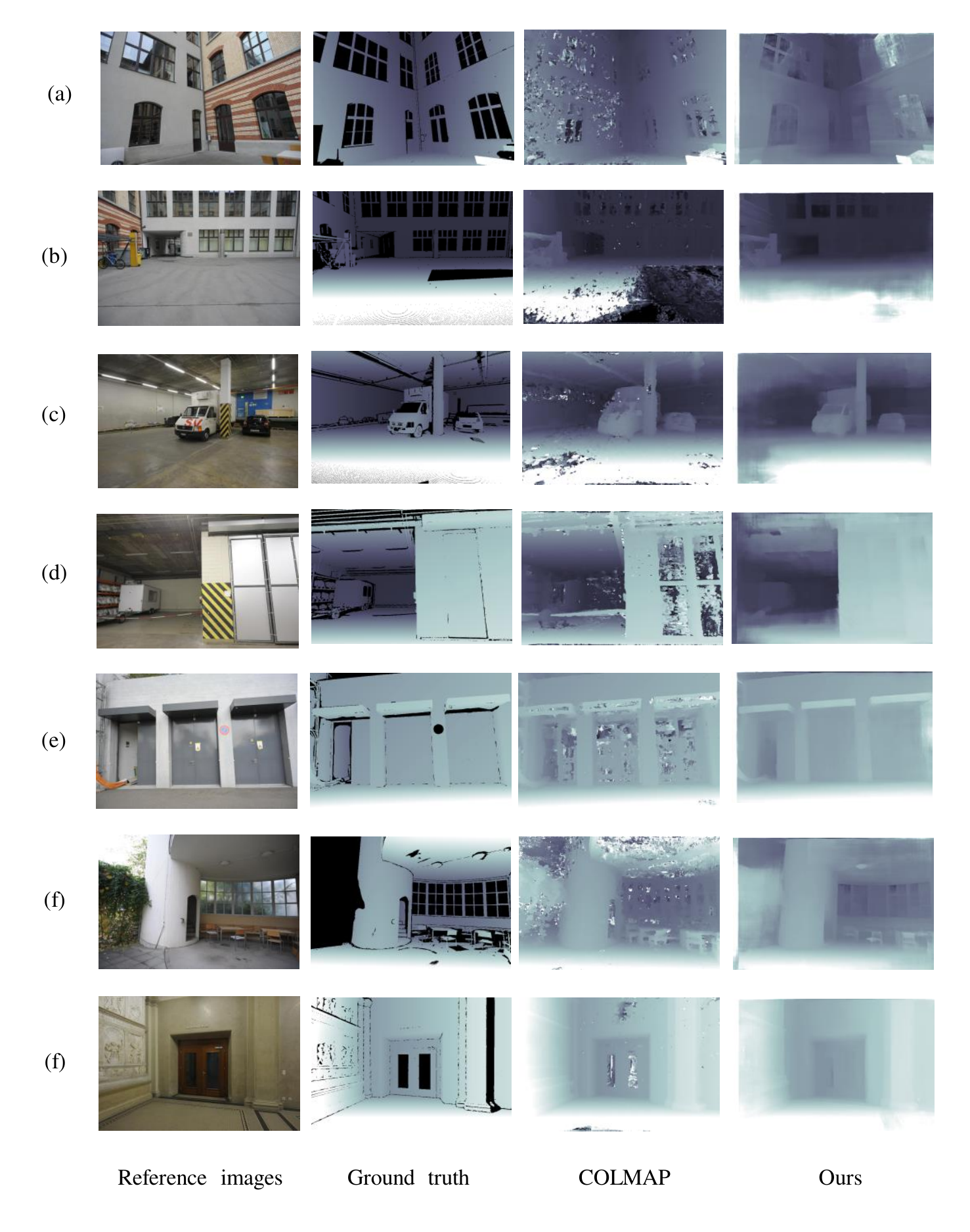}
\caption{Qualitative comparisons with COLMAP \cite{schonberger2016structure} on ETH3D datasets.}
\label{fig:vis_e} 
\end{figure*}

\begin{figure*}[tb]
\centering \includegraphics[width=122mm]{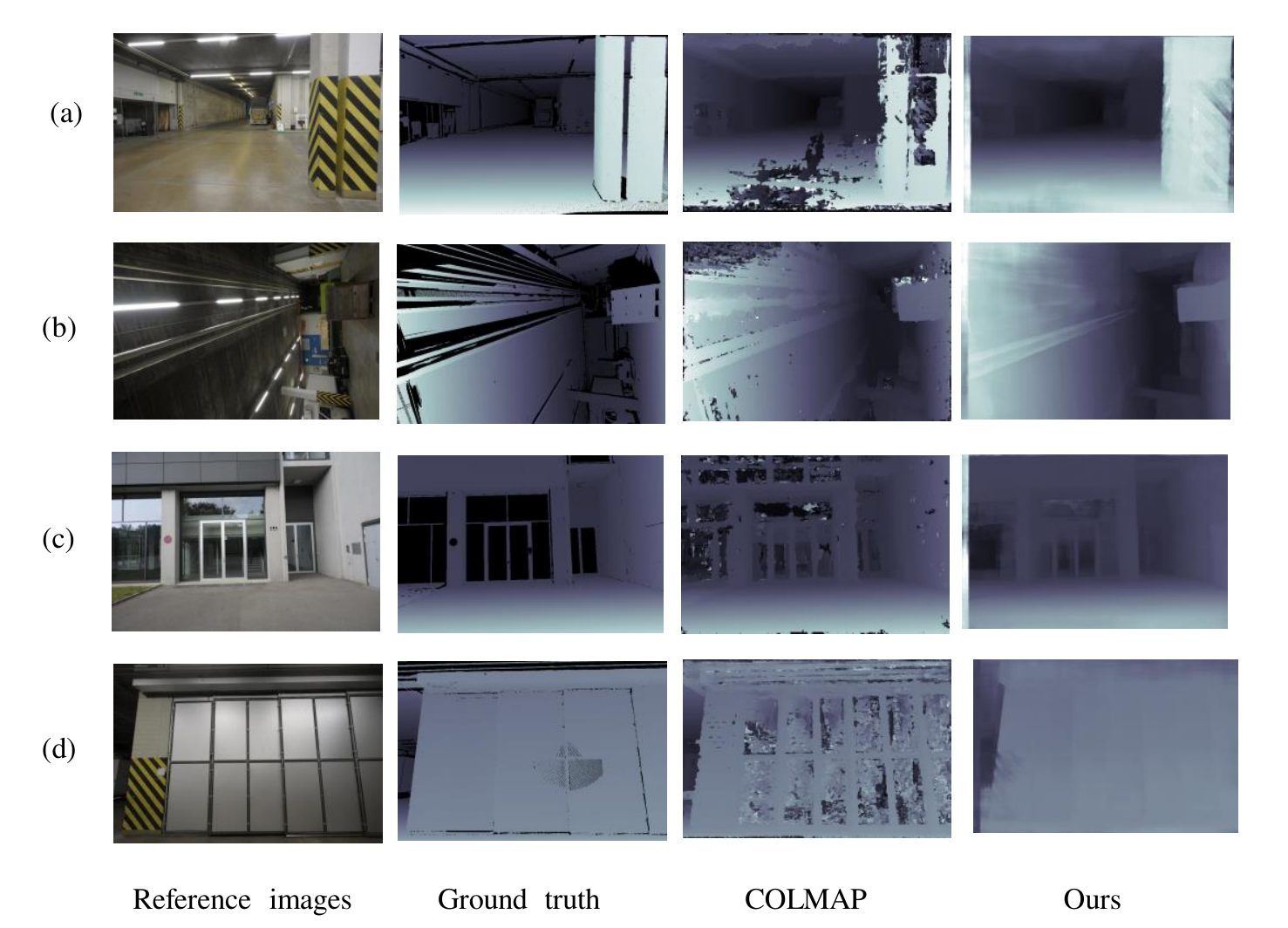}
\caption{Qualitative comparisons with COLMAP \cite{schonberger2016structure} on challenging materials. 
a) Textureless ground and wall. b) Poor illumination scene. c) Reflective and transparent glass wall. d) Reflective and textureless wall.}
\label{fig:vis_e1} 
\end{figure*}

We show some qualitative comparison with the previous methods. Since there are no source code available for BA-Net \cite{tang2018ba}, we compare the visualization results of our method with DeMoN \cite{ummenhofer2017demon} and COLMAP \cite{schonberger2016structure}. Figure \ref{fig:vis_d} shows the predicted dense depth map by our method and DeMoN on the DeMoN datasets. As we can see, demon often miss some details in the scene, such as plants, keyboard and table legs. In contrast, our method reconstructs more shape details. Figure \ref{fig:vis_e} shows some estimated results from COLMAP and our method on the ETH3D dataset. As shown in the figure, the outputs from COLMAP are often incomplete, especially in  textureless area. On the other hand, our method performs better and always produce an integral depth map. 
{In Fig.\ref{fig:vis_e1}, more qualitative comparisons with COLMAP on challenging materials are provided.}


\end{document}